\documentclass[10pt,twocolumn,letterpaper]{article}
\usepackage[pagenumbers]{cvpr}

\usepackage[pagebackref,breaklinks,colorlinks,allcolors=cvprblue]{hyperref}
\usepackage[utf8]{inputenc} 
\usepackage[T1]{fontenc}    
\usepackage{hyperref}       
\usepackage{url}            
\usepackage{booktabs}       
\usepackage{amsfonts}       
\usepackage{nicefrac}       
\usepackage{microtype}      
\usepackage{xcolor}         
\usepackage{graphicx}
\usepackage{amsmath}
\usepackage{makecell}
\usepackage{multirow}
\usepackage{listings}
\usepackage{enumitem}
\usepackage{wrapfig}
\usepackage{tabularx}
\usepackage{amssymb}
\usepackage{needspace}
\usepackage{xr}
\usepackage{cuted}
\usepackage{placeins}

\lstdefinestyle{promptstyle}{
    basicstyle=\ttfamily\scriptsize,
    breaklines=true, 
    breakatwhitespace=true,
    frame=single,
    rulecolor=\color{black!30},
    backgroundcolor=\color{gray!5},
    columns=fullflexible,
    keepspaces=true,
    captionpos=b,
    extendedchars=true,
    showstringspaces=false,
    tabsize=2,
    mathescape=true,
    escapeinside={(*@}{@*)}
}

\definecolor{cvprblue}{rgb}{0.21,0.49,0.74}

\def\confName{CVPR}
\def\confYear{2027}

\title{Beyond End-to-End Black-Box Mapping: An Intentional Agent Framework for Cognitively Driven Facial Reaction Generation}

\author{
\begin{tabular*}{\textwidth}{@{\extracolsep{\fill}}ccc}
\begin{minipage}[t]{0.31\textwidth}
\centering
Hanzhong Zhang\\
Department of Computer Science\\
University of Exeter\\
Exeter, UK\\
{\tt\small armhiabelliard@gmail.com}
\end{minipage}
&
\begin{minipage}[t]{0.31\textwidth}
\centering
Jindong Wang\\
Department of Data Science\\
William \& Mary\\
Williamsburg, VA, USA\\
{\tt\small jdw@wm.edu}
\end{minipage}
&
\begin{minipage}[t]{0.31\textwidth}
\centering
Siyang Song\thanks{Corresponding author: \texttt{s.song@exeter.ac.uk}.}\\
Department of Computer Science\\
University of Exeter\\
Exeter, UK\\
{\tt\small s.song@exeter.ac.uk}
\end{minipage}
\end{tabular*}
}


\begin{document}

\maketitle

\begin{strip}
    \centering
    \includegraphics[width=\textwidth]{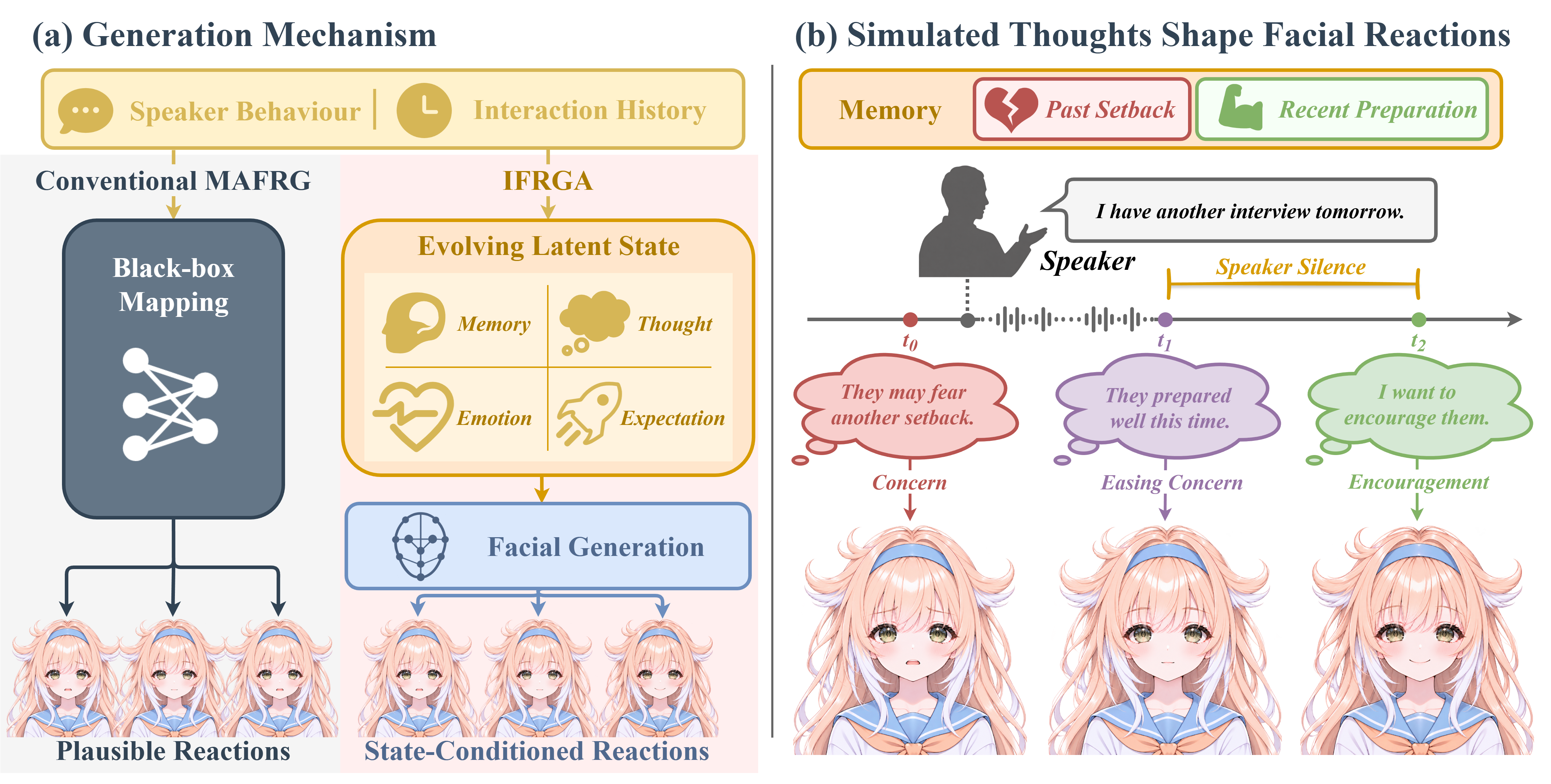}

    \captionsetup{
        hypcap=false,
        skip=1pt
    }
    \captionof{figure}{\textbf{Facial reaction generation through a latent state.} The Intentional Facial Reaction Generation Agent (IFRGA) maintains and updates a latent state for facial reaction generation. \textbf{(a) Generation mechanism.} IFRGA updates the latent state using speaker behaviour and interaction history, then combines simulated thoughts and emotion with speaker behaviour to generate facial reactions. \textbf{(b) Example.} In the illustrated interview scenario, memory of a past setback and recent preparation informs simulated thoughts that evolve from concern to encouragement, guiding facial reactions even during speaker silence.}
    \label{fig:teaser}
    \captionsetup{hypcap=true}
\end{strip}

\begin{abstract}
In dyadic human-human interactions, a human listener's facial reactions can reflect latent states containing memory of earlier interactions, thoughts, emotion, and expectations.
However, many facial reaction generation (FRG) methods predict reactions from speaker behaviour and interaction history without explicitly modelling the listener's thoughts, emotion, and expectations as factors guiding generation.
Furthermore, their generation processes do not explicitly connect ongoing interpretation of the interaction with latent state evolution and facial feedback.
In this paper, we introduce the \textbf{Intentional Facial Reaction Generation Agent} (IFRGA), which uses an explicit latent state to connect the agent's interpretation of the interaction with facial reaction generation.
Specifically, the Latent State Evolution Module (LSEM) uses relevant memory, retained thoughts, and current speaker behaviour to update the latent state through a computational mechanism inspired by human thought processes.
Within LSEM, the Inner Thought Flow (ITF) carries generated thoughts across updates, allowing each interpretation to build on earlier thoughts while incorporating current speaker behaviour.
This mechanism is designed to support real-time, proactive inference during interaction.
Then, the Facial Reaction Generation Module (FRGM) aligns the current thought, emotion values, and available speaker behaviour over time.
It combines predicted facial values with changes from a Diffusion Transformer to generate facial reactions guided jointly by speaker behaviour and latent state evolution.
On REACT 2025, IFRGA obtains a correlation score of 0.8367 and a distance score of 143.64 against reference reactions, both numerically better than published results of compared learnt methods.
A blinded human evaluation demonstrated that carrying thoughts across matched updates significantly improved naturalness and contextual appropriateness ratings compared with carrying factual summaries.
\end{abstract}

\section{Introduction}

Facial Reaction Generation (FRG) is crucial for enabling virtual avatars with anthropomorphic social interaction capabilities \citep{song2023react2023,luo2024reactface,song2023multiple}. In real-world dyadic human-human interactions, a listener's non-verbal feedback (e.g., subtle changes in facial expressions) plays an irreplaceable role in maintaining conversational rhythm, expressing empathy, and conveying interactive intentions \citep{stacchio2025empathic,ng2023can,atzil2023facilitating}. Therefore, how to automatically and adaptively generate contextually appropriate human-like facial reactions from the human user's multi-modal behaviours has become a core research challenge in building highly realistic social virtual humans \citep{song2023react2023,song2025react,luo2024reactface,liang2023unifarn}.

Existing FRG methods typically directly map the perceived speaker behaviours to human-like facial reactions using standard deep learning networks.
Early methods frequently reproduce a single facial sketch of the listener's facial expression from the extracted speaker facial behaviour features, such as facial movement values~\citep{huang2017dyadgan} or estimated emotion~\citep{nojavanasghari2018interactive}, using conditional generative adversarial networks (GANs), where a discriminator assesses whether the generated sketches are real or not.
Since the same speaker behaviour can trigger multiple appropriate facial reactions (AFRs)~\citep{song2023multiple,song2025react}, multiple appropriate facial reaction generation (MAFRG) methods frequently predict a AFR distribution representing multiple different AFRs corresponding to each input speaker behaviour via variational autoencoder (VAE) that predicts a Gaussian distribution from speaker facial and audio features~\citep{song2025react}, Gaussian mixture model (GMM) that combines multiple Gaussian distributions~\citep{nguyen2024multiple}, or reversible graph neural network (GNN) that learns the target distribution from real AFRs to train a speaker behaviour-conditioned AFR generator~\citep{xu2026reversible}. Alternatively, recent diffusion-based MAFRG methods produce multiple AFRs by denoising random values conditioned on the input speaker behaviour. Some of them denoise compressed facial reaction features (e.g., AUs) conditioned on the learnt speaker facial behaviour features~\citep{yu2023leveraging} or earlier speaker facial behaviours~\citep{nguyen2024latent}, while others directly denoise facial reaction features conditioned on currently perceived speaker facial behaviour and audio behaviours~\citep{zhu2024perfrdiff,luo2025reactdiff}. 
In particular, some of these MAFRG methods specifically considered previous speaker behaviours and its previously generated AFRs in producing current AFRs~\citep{luo2024reactface,nguyen2024multiple,zhu2025perreactor}, e.g., constraining changes between previously and existing generated AFR frames~\citep{luo2025reactdiff}.

For a human listener in dyadic interaction, memory of earlier speaker and listener behaviour can influence the listener's thoughts, emotion, expectations, and choices of actions~\citep{scherer2001appraisal,shoda2002personality}. Together with memory, these contents form the latent state considered in this paper, which can change during the interaction~\citep{mischel1995cognitive,shoda2002personality}. Specifically, facial responses can vary with information previously received about another person~\citep{likowski2008attitudes,mermillod2018social} and what people recall or think about~\citep{duken2026reliving,murata2016goal}. They also vary with how people interpret a situation~\citep{ray2010cognitive} and whether information about another person's behaviour meets their expectations~\citep{bartholow2001expectancy}. These findings motivate our hypothesis that explicitly maintaining and updating a latent state may help FRG generators produce more appropriate facial reactions. By contrast, the FRG methods discussed above do not explicitly model how the listener's thoughts, emotion, and expectations jointly guide generation~\citep{huang2017dyadgan,xu2026reversible,zhu2025perreactor,zhu2024perfrdiff,luo2024reactface,luo2025reactdiff} (\textbf{Problem 1}).

Furthermore, these FRG methods organise generation around predicting facial reactions from speaker behaviour, with historical behaviour and previously generated reactions used to model temporal relationships~\citep{luo2024reactface,luo2025reactdiff}. Human facial behaviour, however, can also be shaped by a person's own goals, interpretation, and regulation of the response~\citep{murata2016goal,ray2010cognitive,gross1993emotional}. This motivates modelling how a listener continually interprets incoming information, so that facial behaviour reflects both perceived speaker behaviour and the listener's ongoing thought processes. Yet, the FRG methods discussed above lack an explicit mechanism for proactively forming and revising this interpretation in real time and combining the resulting latent state changes with speaker behaviour to generate facial reactions (\textbf{Problem 2}).

In this paper, we introduce the Intentional Facial Reaction Generation Agent (IFRGA), an agent architecture that combines speaker behaviour with an evolving latent state for FRG. To address Problem 1, IFRGA maintains memory, generated thoughts, emotion, expectations, and action-related information in an explicit latent state. To address Problem 2, we design a computational mechanism inspired by human thought processes to support real-time, proactive inference and facial feedback. Specifically, the Latent State Evolution Module (LSEM) uses relevant memory, the retained latent state, and current speaker behaviour to update the agent's interpretation of the interaction. At each update, a large language model (LLM)~\citep{park2023generative,xi2025rise} generates a thought describing this interpretation, forms an expectation, and decides whether to produce an action. The Inner Thought Flow (ITF) carries generated thoughts across updates, so that each new thought builds on earlier thoughts while incorporating current speaker behaviour. LSEM also updates emotion using the available information and retained emotion. The Facial Reaction Generation Module (FRGM) then aligns the current thought in the ITF, emotion values, and available speaker behaviour over time to generate facial reactions. Consequently, facial reactions are guided jointly by speaker behaviour and the thought and emotion produced through latent state evolution. In summary, our main contributions are as follows:

\begin{itemize}[leftmargin=2em,itemsep=0pt,topsep=2pt,parsep=0pt,partopsep=0pt]

\item We introduce an agent architecture for FRG that explicitly represents memory, generated thoughts, emotion, and expectations in a latent state. This provides a maintained basis for interpreting the interaction and allows facial reactions to depend on the agent's current latent state together with speaker behaviour.

\item We develop a latent state evolution mechanism, inspired by human thought processes, to support real-time, proactive inference for FRG. LSEM combines current speaker behaviour with memory and retained thoughts to form and revise the agent's interpretation throughout the interaction. FRGM then uses the current thought and emotion together with speaker behaviour to generate facial reactions, combining predicted facial values with changes generated by a Diffusion Transformer (DiT).

\item On the REACT 2025 test set, IFRGA obtains an FRCorr of 0.8367 and an FRDist of 143.64. In separate human evaluations, the full method receives a significantly higher mean score than the event-triggered and heuristic baselines, while carrying thoughts across updates receives significantly higher ratings of contextual appropriateness and naturalness than retaining factual summaries under matched update schedules and token budgets.

\end{itemize}

\section{Related Work}

\textbf{Facial Reaction Generation.}
Early work used speaker cues to predict listener behaviour or generate facial reactions \citep{Cassell01051999,morency2008predicting,huang2017dyadgan,nojavanasghari2018interactive,song2022learning}. Subsequent work drew on general image synthesis and Talking Head generation to improve visual fidelity \citep{karras2019style,thies2016face2face,vougioukas2020realistic,zhang2023sadtalker}, while Multiple Appropriate Facial Reaction Generation (MAFRG) addressed the inherent one-to-many nature of listener responses \citep{song2023multiple,song2023react2023,song2024react,song2025react}. Existing MAFRG methods include Transformer and contrastive-learning approaches \citep{liang2023unifarn,hoque2023beamer}, online synchronous and VAE-based latent-variable models \citep{luo2024reactface,liu2024one,nguyen2024multiple}, and diffusion frameworks \citep{yu2023leveraging,nguyen2024vector,nguyen2024latent,luo2025reactdiff}. These diffusion-based and generative architectures have been continuously refined to enhance context-awareness and temporal coherence through techniques such as multi-scale frequency scattering transforms \citep{mao2025scattering}, hierarchical 2D--3D multimodal decoupling and cognitive disentanglement \citep{lv2025hierarchical,luo2026mreactor}, motion-aware visual dynamic fusions \citep{huang2025multiple}, and speaker-anchored residual denoising \citep{liu2026resdifffrg}.

Some MAFRG methods learn appropriate reaction distributions via Graph Neural Networks \citep{xu2026reversible}. To account for differences between listeners responding to the same speaker behaviour, other methods edit generic network weights or disentangle listener styles to mimic personalised cognitive styles \citep{zhu2024perfrdiff,zhu2025perreactor}, or condition diffusion generation on explicit Big Five personality traits \citep{wang2025explaining}. However, these methods do not explicitly describe how to maintain and update a latent state containing generated thoughts, emotion, and expectations during an interaction. Our IFRGA maintains and updates such a latent state and uses the current thought and emotion together with available speaker behaviour to generate facial reactions.

\textbf{Proactive and Cognitive Agents.}
LLM-based agents extend passive instruction following towards proactive interaction through perception, planning, memory, and internal reasoning \citep{deng2025proactive,luo2025large,lu2024proactive}. Embodied agents further connect language models with environmental perception and action \citep{wang2023voyager,huang2022inner}, with proactive mechanisms applied to visual analytics \citep{zhao2025proactiveva} and programming assistance \citep{zhao2025codinggenie}. In social interaction, inner-thought mechanisms can determine intervention timing \citep{liu2025proactive}, while personalised modelling supports longer-term user adaptation \citep{qiu2025measuring,garbacea2025hyperalign}. Some of these systems generate thoughts during conversational pauses to decide when to participate \citep{liu2025proactive}. Building on this direction, IFRGA uses memory of earlier speaker behaviour and its own prior actions, retained thoughts, and current speaker behaviour to form and revise its interpretation of the interaction. FRGM then uses the resulting thought and emotion together with speaker behaviour to generate facial reactions.

\section{Intentional Facial Reaction Generation Agent}

\begin{figure}[t!]
  \centering
    \includegraphics[width=\linewidth]{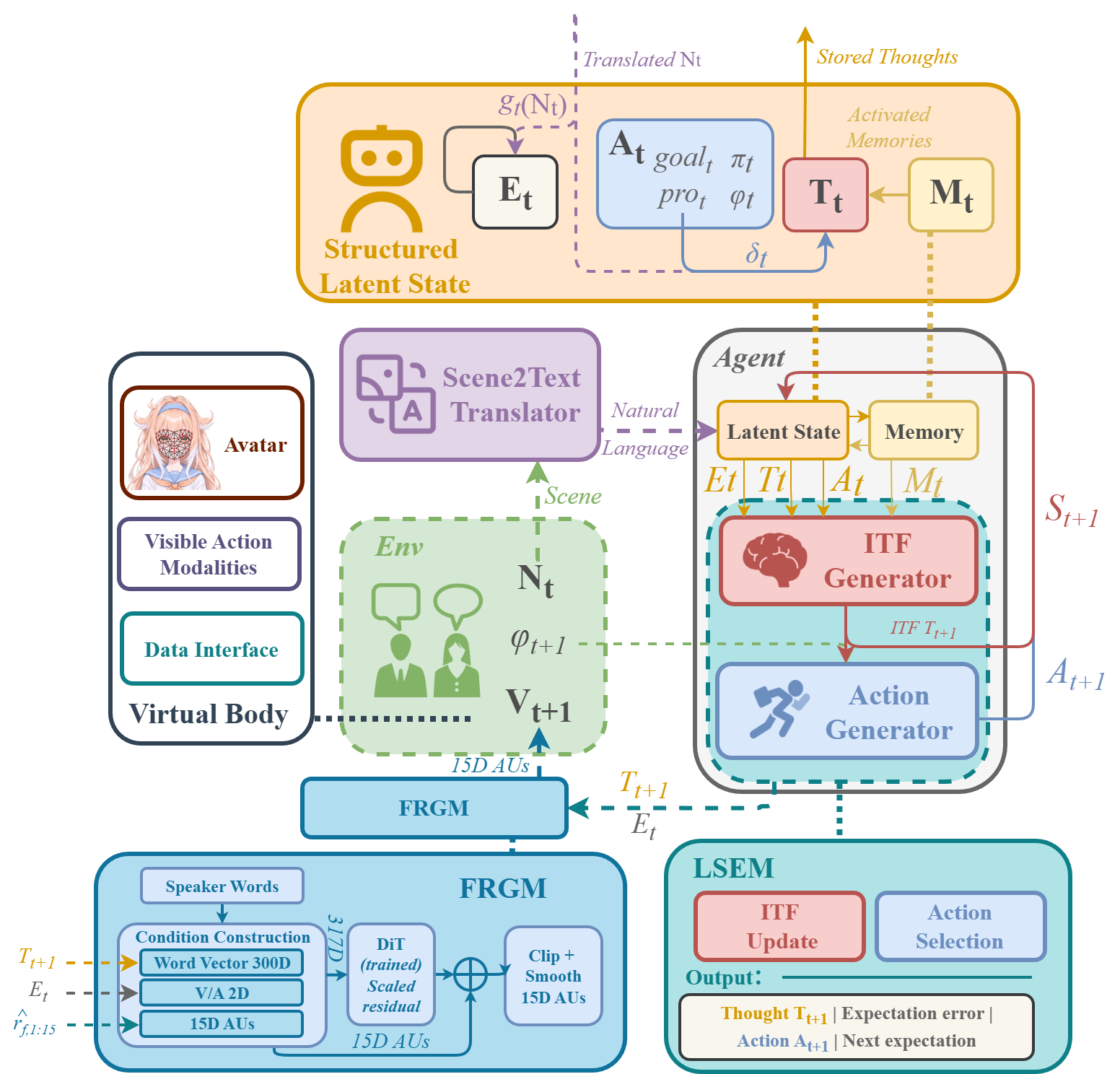}
    \caption{\textbf{Overview of IFRGA.} LSEM updates the latent state, including the ITF, using one LLM call and programmed updates. FRGM uses the ITF, emotion in the latent state, and speaker behaviour to generate facial reactions, even when no action is produced. The DiT receives 317 values: a 300-dimensional vector from the ITF and available speaker words, valence and arousal from the latent state, and 15 AU values from the 25-dimensional vector $\hat{\mathbf r}_f$ predicted for frame $f$. The DiT produces 25 changes, which are combined with $\hat{\mathbf r}_f$; the first 15 resulting values are used as AUs to render the facial reaction.}
    \label{fig:overall_architecture}
\end{figure}

\subsection{Overview}
\label{sec:overview}

This section presents our \textbf{Intentional Facial Reaction Generation Agent} (IFRGA) for the FRG task. IFRGA uses a latent state evolution mechanism inspired by human thought processes to support proactive inference, while FRGM combines the resulting thought and emotion with speaker behaviour to generate facial reactions.

As illustrated in Fig.~\ref{fig:overall_architecture}, our IFRGA is made up of three steps: selecting relevant memory, updating the latent state while selecting an action if needed, and generating facial reactions. The first two steps are performed by LSEM, while the third step is performed by FRGM.
Given the current speaker behaviour described by $N_t$, previous thought $T_t$, emotion $E_t$ and information about actions $A_t$, LSEM first selects relevant memory $M_t$ as:
\begin{equation}
    M_t = \operatorname{Select}(\mathcal{M}_t, N_t, T_t),
\end{equation}
where $\mathcal{M}_t$ is the memory stored by IFRGA, and $M_t$ is the memory selected for the current update. $N_t$ is the text description of the speaker behaviour, while $T_t$ is the retained thought and $E_t$ contains the retained values representing emotion. $A_t$ denotes retained action-related information. The implementation records the selected action text, its semantic expectation, and the reported prediction error. Purpose and feasibility are expressed through the thought and action descriptions. In the online agent, memory selection uses $N_t$ and $T_t$. For the offline REACT evaluation, a training example is selected using the speaker's full transcript, and recent states are included in the memory at each update. This allows IFRGA to use information from earlier interactions when updating the latent state.

Then, LSEM further uses $M_t$ to update the latent state and generate an action as:
\begin{equation}
    (S_{t+1}, \mathrm{at}_{t+1}) = \mathcal{U}(S_t, N_t),
    \qquad
    S_t = \{M_t, T_t, E_t, A_t\},
\end{equation}
where $S_t$ is the latent state supplied to the current update after memory selection, and $S_{t+1}$ is the updated latent state. $\mathcal{U}$ denotes the complete update, which includes one LLM call and programmed updates to memory and emotion. The LLM call jointly produces the next thought, an action, the next expectation, and a prediction error. The prediction error is a value indicating how much the current information conflicts with the previous expectation. $\mathrm{at}_{t+1}$ denotes the action text if an action is produced; otherwise, it denotes the absence of an action. As a result, IFRGA retains the updated latent state for the next update. Each subsequent update combines current speaker behaviour with the retained latent state, allowing IFRGA's interpretation and emotion to develop throughout the interaction.

Finally, FRGM uses the ITF, the emotion in the latent state, and the available speaker behaviour to generate facial reactions. First, the current thought in the ITF is converted into a 300-dimensional vector. When a word from the speaker is available, the vector of that word is averaged with the ITF vector over the corresponding time interval. FRGM obtains valence and arousal from the latent state. A trained model uses the speaker's facial attributes, audio and text, together with valence and arousal, to predict 25 values describing facial movements. The first 15 values correspond to Action Units (AUs). The 300-dimensional vector, valence, arousal, and these 15 AU values form a 317-dimensional condition for a trained Diffusion Transformer (DiT). The DiT predicts changes to the 25 values, which are added to the values predicted by the trained model. The results are smoothed and limited to valid ranges, and the first 15 AU values are used to render the facial reaction. FRGM can therefore generate facial reactions even when no action is produced.

\subsection{Latent State Evolution Module}
\label{sec:lsem}

LSEM implements a latent state evolution mechanism inspired by human thought processes to support proactive inference during interaction. Specifically, it combines current speaker behaviour with memory and the retained thought, emotion, and expectation to update IFRGA's interpretation of the interaction. LSEM first selects relevant memory and then updates the latent state, including the ITF, whose current thought is supplied to FRGM together with emotion and available speaker behaviour.

LSEM first prepares $M_t$ so that information from earlier interactions is available for updating the latent state. In online interaction, LSEM combines the text description of the current speaker behaviour $N_t$ with the retained thought $T_t$ in the ITF and converts the combined text into a vector. LSEM compares this vector with the vectors of stored memories $\mathcal{M}_t$ to select relevant memories. The selected memories are combined with the most recent memories to form $M_t$. For the offline REACT evaluation, a training example is selected using the speaker's full transcript, while up to three recent states are included in $M_t$ at each update. The resulting $M_t$ becomes part of the latent state and is used with $N_t$ and $T_t$ in the current update.

With $M_t$, LSEM updates the emotion in the latent state before calling the LLM. It obtains new values from the current speaker behaviour $N_t$, the selected memory $M_t$, and the retained thought $T_t$, then combines the new values with the previously retained emotion. The prediction error from the preceding update affects both the new values and the amount of the previous emotion retained. LSEM then provides $N_t$, $M_t$, the updated emotion $E_t$, $T_t$, and the previous expectation, which is represented in the action vector $A_t$, to the LLM in a single call. The LLM produces the next thought $T_{t+1}$, decides whether to produce an action, and produces the next expectation and a new prediction error. $T_{t+1}$ continues the ITF and is retained with the next expectation and the new prediction error for the next update. Here, $T_{t+1}$ is generated text expressing IFRGA's current interpretation of the interaction. Retaining this thought allows subsequent inference to build on earlier interpretations while incorporating current speaker behaviour and relevant memory. The state evolution and its connection to facial generation are detailed in Appendix~\ref{app:eq_validation}, while Appendix~\ref{app:innerthought_prompt} provides the update prompt. To generate facial reactions, FRGM uses $T_{t+1}$ together with the updated emotion $E_t$ and the available speaker behaviour, even when no action is produced.

\subsection{Facial Reaction Generation Module}
\label{sec:frgm}

LSEM updates the ITF and the emotion in the latent state, but FRG also requires facial reactions that reflect these updates. Facial Reaction Generation Module (FRGM) therefore uses the ITF, the emotion in the latent state, and the available speaker behaviour to generate facial reactions, even when no action is produced. FRGM first converts the current thought in the ITF and available speaker words into a vector and predicts 25 values for a facial reaction from the emotion and speaker behaviour; it then uses the vector to condition a DiT and combines the DiT's output with the 25 values to generate the facial reaction.

FRGM first converts the current thought $T_{t+1}$ in the ITF into a 300-dimensional vector. When a speaker word is available, FRGM averages the vector of that word with the 300-dimensional vector over the word's time interval; otherwise, the 300-dimensional vector remains unchanged. FRGM also obtains valence and arousal from the emotion $E_t$ in the latent state: valence describes how positive or negative the emotion is, whereas arousal describes its level of activation. Both values are aligned with the same time axis and scaled for the DiT.

A trained model predicts a 25-dimensional vector $\hat{\mathbf r}_f$ for each frame $f$ from the speaker's facial attributes, numbers extracted from the speaker's audio and transcript, and valence and arousal from $E_t$. $\hat{\mathbf r}_{f,1:15}$ contains the AU values, $\hat{\mathbf r}_{f,16:17}$ contains the predicted valence and arousal of the facial reaction, and $\hat{\mathbf r}_{f,18:25}$ contains the predicted probabilities of eight emotions. FRGM combines the 300-dimensional vector, valence and arousal from $E_t$, and $\hat{\mathbf r}_{f,1:15}$ to form the 317 values used to condition the DiT. The complete $\hat{\mathbf r}_f$ is retained for combination with the DiT output in the next step. The DiT uses the 317 values from the previous step to gradually transform randomly generated values into 25 changes for each frame $f$, one for each value in $\hat{\mathbf r}_f$. Before combining them with $\hat{\mathbf r}_f$, FRGM reduces slow changes in the DiT output and limits their size using statistics from the training data. FRGM then adds the adjusted changes to $\hat{\mathbf r}_f$ and smooths the result over time.

The main numerical protocol applies a stochastic AU-probit readout with coefficient .55, specified in Appendix~\ref{app:emotion_mapping}. The complete 25-channel sequence enters numerical evaluation, and its first 15 AU channels are supplied to the avatar adaptor.

\section{Experiments}

\subsection{Settings}

\textbf{Dataset.} We evaluate IFRGA on the MARS dataset from the REACT 2025 challenge \citep{song2025react}. The dataset contains 137 recorded dyadic interactions involving 23 speakers and 137 listeners. These interactions are divided into 2,856 pairs of multimodal sessions, with each pair containing a speaker's and a listener's recordings. The original recordings contain synchronised audio and facial videos and last from 19 minutes and 50 seconds to 40 minutes and 49 seconds. We retain natural speech gaps to evaluate reactions under discontinuous conversational timing. Speaker facial observations remain available in the natural-gap setting. In controlled probes, we specify what information is provided at each update. In fixed-no-change probes, no new utterances, audio features, video frames, or visual descriptions are supplied after initialisation; subsequent updates receive a fixed message indicating no change. In the listener-task silence ablation, updates follow preset time steps without new speaker events; conditions retaining perception receive simulated feedback indicating whether the agent's action is ongoing or complete. The corresponding settings are detailed in Appendices~\ref{app:matched_itf_control} and~\ref{app:Controlled_Empirical_Validation}.

We additionally test recording-group-disjoint reaction generation with an Action-conditioned component configuration. The component models use 78 fitting groups and 12 validation groups from the official training partition. Confirmation covers 242 eligible windows from 14 official test recording groups absent from the official training and validation partitions. Full obtains VaG .03964, VeG .01924, and DG .10315, with lower gaps than Learnt-Only and Mapping-Only on all three measures. All six recording-level comparisons have Holm-adjusted $p\leq .003052$. The empirical terminal initialisation, three training seeds, and ten actual facial samples per input are specified in Appendix~\ref{app:recording_generalisation}.

We further evaluate modular adaptation on \textit{Seamless Interaction}~\citep{seamless_interaction}. Keeping the state-update model, prompts, and low-frequency mapping fixed, we adapt only the facial DiT using conditions produced by the inference pipeline. Under a common mean-preserving decoder, AU-MAE decreases from .23034 to .10057 and the velocity gap decreases from .00948 to .00460. Both paired comparisons have $p_{\mathrm{Holm}}=.00098$. The protocol and controls are reported in Appendix~\ref{app:seamless_adaptation}.

\textbf{Implementation Details.} The retained main benchmark configuration uses frozen Qwen3-8B Q4\_K\_M with an 8192-token context and four-second observation bins. Training-set interactions populate the agent's memory. Inference uses speaker observations and training-set memories, while target listener reactions enter evaluation. Each directed input shares one ITF trajectory across ten independently sampled facial predictions. Facial generation uses 20 DDIM steps and CFG=3. A learnt 72-dimensional condition mapping supplies low-frequency facial values, and the DiT predicts a 25-dimensional residual. The recording-group-disjoint component study uses its separately trained Action/State configuration and empirical terminal initialisation, specified in Appendix~\ref{app:recording_generalisation}. The affect-estimator and diversity components follow their recorded conditions in Appendices~\ref{app:vad_replacement} and~\ref{app:diversity_audit}.

Agent affect is represented in VAD space~\citep{russell1977evidence}. The NRC VAD Lexicon v2~\citep{mohammad2025nrc} provides text-derived affective values. On the official EmoBank test set~\citep{buechel-hahn-2017-emobank}, its Pearson correlations for Valence, Arousal, and Dominance are .512, .236, and .151. Details are provided in Appendix~\ref{app:emotion_mapping}.

We evaluate affect-estimator replacement using a separately trained Action-conditioned facial component. Generated Action text, word semantics, facial weights, paired sampling, and post-processing are fixed while NRC V/A readout is replaced with a regressor trained on human EmoBank annotations. On 14 official validation recordings and 225 windows, AU-MAE decreases from .257147 to .249118, low-frequency AU-MAE from .217417 to .209793, and jerk from .013391 to .012258. The corresponding Holm-adjusted $p$ values across five trajectory metrics are .03418, .04980, and .000610. A separate 12-recording confirmation also supports these three improvements. Complete metrics and controls are reported in Appendix~\ref{app:vad_replacement}.

\textbf{Interactive Feasibility.} On an RTX 5090 Laptop GPU, 24 complete updates using the formal 20-step configuration take 1.9645s on average and 2.3510s at the 95th percentile. Timing begins at perceived-text availability and ends at submission in headless Chromium using SwiftShader. Each display update generates one 200-frame trajectory. The 3-step development configuration takes 1.7936s on average. Incoming events trigger cancellation requests in 34.71ms on average, and an event received during idle waiting starts state inference after 133.29ms. Measurement boundaries and scheduling probes are reported in Appendix~\ref{app:runtime_comparison}.

\textbf{Metrics.} We report the standard REACT 2025 appropriateness, diversity, and synchrony metrics. The human study evaluates five perceptual dimensions and reports results from the Reaction Quality Scorer (RQS). Additional random-context studies directly assess naturalness and contextual appropriateness on 1--7 scales, with contextual appropriateness as the primary outcome. Paired human comparisons are described in Section~\ref{sec:additional_perception}. RQS provides a supplementary AU-compatibility measure, and LLM-as-a-Judge scores provide action-level diagnostics. Definitions and protocols are reported in Appendices~\ref{app:exp_setting} and~\ref{app:rqs_details}.

\subsection{Comparison with existing MAFRG methods}

Table~\ref{tab:performance_comparison} reports evaluation on both role directions of the 625 test interaction pairs, yielding 1,250 directed inputs with ten generated reactions per input. IFRGA obtains an FRCorr of .8367 and an FRDist of 143.64, both numerically better than the listed learnt reference values. Its FRDiv, FRVar, and FRSyn are .0181, .0253, and 45.38, respectively. The reference results are taken from the cited publications and were not re-evaluated under our setup.

Because FRG is a one-to-many task, neither exact ground-truth correlation nor diversity alone provides a complete quality measure. We therefore report the standard metrics together with human evaluation, the controlled diversity audit in Appendix~\ref{app:diversity_audit}, and the qualitative comparisons in Figure~\ref{fig:qualitative_frames}.

\begin{table*}[htbp]
  \centering
  \caption{\textbf{Quantitative comparison on the REACT 2025 dataset.} Our results use both role directions of 625 test interaction pairs, yielding 1,250 directed inputs with ten generated reactions per input. For IFRGA, FRSyn is computed from the first generated reaction. Other methods use published reference results and were not re-evaluated under our setup. Bold indicates the best numerical result among the listed learnt methods. Arrows indicate the preferred direction of each metric.}
  \label{tab:performance_comparison}
  \begin{tabular}{lccccc}
    \toprule
    \multirow{2}{*}{Method} & \multicolumn{2}{c}{Appropriateness} & \multicolumn{2}{c}{Diversity} & Synchrony \\
    \cmidrule(lr){2-3} \cmidrule(lr){4-5} \cmidrule(lr){6-6}
     & FRCorr ($\uparrow$) & FRDist ($\downarrow$) & FRDiv ($\uparrow$) & FRVar ($\uparrow$) & FRSyn ($\downarrow$) \\
    \midrule
    GT \citep{song2025react} & 10.00 & 0.00 & 0.1876 & 0.0669 & 48.66 \\
    B\_Random \citep{song2025react} & 0.03 & 474.68 & 0.3342 & 0.1671 & 46.64 \\
    B\_Mime \citep{song2025react} & 0.52 & 206.02 & 0.0000 & 0.0766 & 43.70 \\
    B\_MeanFr \citep{song2025react} & 0.00 & 205.65 & 0.0000 & 0.0000 & 49.00 \\
    Trans-VAE \citep{song2025react} & 0.30 & 181.72 & 0.0076 & 0.0083 & 49.00 \\
    ReGNN \citep{xu2026reversible,song2025react} & 0.54 & 155.49 & 0.0012 & 0.0048 & \textbf{44.81} \\
    PerFRDiff \citep{zhu2024perfrdiff,song2025react} & 0.58 & 217.32 & \textbf{0.1596} & \textbf{0.0862} & 48.66 \\
    SC-Diff \citep{mao2025scattering} & 0.59 & 214.67 & 0.0400 & 0.0811 & 49.00 \\
    \cite{huang2025multiple} & 0.67 & 178.27 & 0.1510 & 0.0801 & 48.07 \\
    \cite{wang2025explaining} & 0.71 & 173.01 & 0.1405 & 0.0769 & 47.77 \\
    \textbf{IFRGA} & \textbf{0.8367} & \textbf{143.64} & 0.0181 & 0.0253 & 45.38 \\
    \bottomrule
  \end{tabular}
\end{table*}

\subsection{Controlled Perceptual Evaluation}
\label{sec:additional_perception}

We conducted two additional studies to evaluate the contextual appropriateness and naturalness of facial reactions, using contexts from pre-existing random selections. In the first study, six participants each rated 48 clips from 12 contexts, yielding 288 ratings for four conditions: Full (orig.), B\_Mime, Full (comp.), and Learnt-Only. In the second study, four participants each rated 24 clips, yielding 96 ratings for Full (ITF) versus Factual Summary on 24 contexts and Full (orig.) versus CaReDiff-face on 12 contexts. Among the 36 contexts in the second study, 24 were rated by one participant and 12 by two. Each paired comparison used the same avatar, camera, and AU display amplitude, with additional idle animation disabled. Participants rated contextual appropriateness and naturalness on 1--7 scales without seeing the method labels.

Full (orig.) uses the original facial-generation implementation, whereas B\_Mime uses the speaker's facial attributes as the predicted reaction. Full (comp.) combines low-frequency facial values with a learnt residual, whereas Learnt-Only directly predicts facial values without the low-frequency mapping; the two are separately trained under matched settings. Full (ITF) carries the preceding thought across updates, whereas Factual Summary carries a summary of observed facts. CaReDiff-face uses published generic offline weights and real 25-dimensional speaker facial attributes; audio and three-dimensional morphable model (3DMM) coefficients are not supplied and are handled by the model's native missing-modality path~\citep{ivpl2026carediff}. For each participant and context, we compute the score difference between the two conditions, then average these differences across participants within each context. We apply a two-sided Wilcoxon signed-rank test, a paired test of the resulting context-level differences. Each study contains two comparisons and two outcomes, so its four $p$ values are corrected using Holm's procedure. As shown in Table~\ref{tab:additional_human}, all four comparisons have positive mean paired differences for both ratings (all $p_{\mathrm{Holm}}<.004$). Full protocols and confidence intervals are reported in Appendix~\ref{app:additional_human_protocol}.

\begin{table}[t]
\centering
\small
\setlength{\tabcolsep}{3pt}
\renewcommand{\tabularxcolumn}[1]{m{#1}}
\caption{\textbf{Additional blinded human comparisons.} $\Delta_{\mathrm{ctx}}$ and $\Delta_{\mathrm{nat}}$ denote mean paired differences in contextual appropriateness and naturalness, respectively; positive values favour the first condition. $p_{\mathrm{ctx}}$ and $p_{\mathrm{nat}}$ are the corresponding Holm-corrected $p$ values. The first two rows use the six-participant study, and the final two use the four-participant study. Holm correction is applied separately within each study.}
\label{tab:additional_human}

\begin{tabularx}{\columnwidth}{
    @{}
    >{\raggedright\arraybackslash}X
    rrrrr
    @{}
}
\toprule
Comparison & Contexts
& $\Delta_{\mathrm{ctx}}$ & $\Delta_{\mathrm{nat}}$
& $p_{\mathrm{ctx}}$ & $p_{\mathrm{nat}}$ \\
\midrule
Full (orig.) -- B\_Mime
& 12 & +1.792 & +1.861 & .001953 & .001953 \\

Full (comp.) -- Learnt-Only
& 12 & +.972 & +.583 & .001953 & .003906 \\

Full (ITF) -- Factual Summary
& 24 & +.854 & +.958 & .001057 & .000240 \\

Full (orig.) -- CaReDiff-face
& 12 & +1.458 & +1.042 & .003906 & .003906 \\
\bottomrule
\end{tabularx}
\end{table}

\paragraph{Matched-update ITF control.}
For the human comparison, both conditions use Qwen3.5-9B. They receive the same observation history, initial state, prompts, logical update times at 0, 2, 4, and 6 seconds, and per-call caps of 256 output tokens and 8,192 context tokens. Full (ITF) carries the preceding thought, whereas Factual Summary carries an observation-based factual summary; both continue to update emotion, expectation, and action. Thus, the higher ratings for Full (ITF) support a benefit of carrying the preceding thought under these matched settings. The complete intervention and seed protocol is given in Appendix~\ref{app:matched_itf_control}.

A separate automatic test uses Qwen3-8B and 14 recording contexts with matched update times and a 180-token output cap. Full achieves higher intention retention than factual-summary carry. A control that reintroduces the initial intention at every update obtains the highest intention margin. The complete three-condition protocol and comparisons are reported in Appendix~\ref{app:matched_itf_control}.

\subsection{Human and Automatic Evaluation}

\paragraph{Human Evaluation.}
\label{sec:rqs_human}
Eight blinded raters each evaluated 96 facial reactions from four conditions in 24 contexts. These contexts came from 14 recording groups. The raters assessed naturalness, contextual appropriateness, emotion fit, temporal continuity, and credibility on 1--7 scales. Event-Triggered used Full's output when speaker input was present and held the last AU frame during speech gaps, while Heuristic-Only used a rule-based mapping from valence and arousal to AU values without the DiT residual. Across the five rating criteria, Full obtained a mean score of 5.527, higher than Event-Triggered (3.059) and Heuristic-Only (1.445), both $p_{\mathrm{Holm}}<.001$ using contexts as paired units. Full human-evaluation details and results for all four conditions are reported in Appendix~\ref{app:rqs_human}.

\paragraph{Reaction Quality Scorer.} Previous FRG work has used speaker--listener behaviour pairs to learn their relationship~\citep{hoque2023beamer}, while other work has trained a discriminator to assess whether a listener's facial reaction is appropriate for the speaker's behaviour~\citep{zhu2025perreactor}. In dialogue evaluation, a scoring model has also been trained on observed and randomly paired utterances to assess how well a response fits the preceding utterance~\citep{tao2018ruber}. Similarly, we trained a Reaction Quality Scorer (RQS) to provide a supplementary automatic estimate of compatibility between the speaker's facial behaviour and the listener's facial reaction using their 15-dimensional AU sequences. RQS was trained on recorded speaker--listener pairs and their perturbed versions, without using facial reactions generated by IFRGA. Across the 96 reactions, RQS scores correlated with mean human ratings (Pearson $r=.855$; Spearman $\rho=.821$, both $p<.05$ at the stimulus level). After retraining RQS without the 14 recording groups represented in the human evaluation, the corresponding correlations were $r=.744$ and $\rho=.788$ on the same reactions. The RQS architecture, training, session-level scores, and further human-correlation analyses are reported in Appendices~\ref{app:rqs_details}, \ref{app:rqs_session_results}, \ref{app:rqs_group_disjoint}, and \ref{app:rqs_metric_comparison}.

\paragraph{Action-level diagnostics.}
We additionally evaluate generated actions across foundation models using five LLM-as-a-Judge criteria. The results and evaluation protocol are reported in Appendix~\ref{app:action_diagnostics}.

\subsection{Ablation Studies}
\label{sec:ablation}

We evaluate the facial component and the retained state separately. To examine the facial component, we compare Full, which combines predicted facial values with changes generated by a neural network, with Learnt-Only, which is trained separately without the mapping used to predict those facial values. To measure the difference between generated and real facial movements, we smooth their AU sequences and calculate the mean absolute difference between them; a lower value indicates a smaller difference in gradual facial changes. Across 60 contexts, Full obtains an error of .23403, compared with .25143 for Learnt-Only, and has a lower error in 39 contexts ($p=.01812$). During long silences in 57 valid contexts, the corresponding errors are .22728 and .27599 ($p=.00905$).

To examine the retained state during speaker silence, we use 16 contexts, each with three language seeds and updates at simulated times of 1, 2, and 3 seconds after silence begins. Full retains perception, memory, emotion, thought, and expectation, whereas each Only condition retains only its named direct input. Retained memory can still contain earlier thoughts, actions, and expectations. A simulated executor tracks the action issued at the start of silence and provides execution feedback only to conditions retaining perception. Full, Only-Perception, and Only-Emotion produce no new action commands, while the proportion of updates producing a new command is .12500 for Only-Thought ($p_{\mathrm{Holm}}=.01262$ for Full versus Only-Thought). The mean self-reported prediction error, a value provided by the model, is .05556 for Full, compared with .13785, .45069, and .57222 for Only-Emotion, Only-Thought, and Only-Expectation, respectively. The adjusted $p$ values for these three comparisons are .03871, .00392, and .00304, respectively, with seven comparisons corrected together.

To examine whether the retained state helps maintain information during continued thought updates, we compare Full, No-Memory, and Only-Thought in 24 newly constructed contexts without new speaker information. All conditions start with the same thought containing the complete initial facts and perform 80 updates over 30 minutes of simulated time, with three language seeds per context. For each thought, we calculate the fraction of four predefined groups of words or phrases related to the initial facts that appear in the thought. The mean fractions are 20.89\% for Full, 3.08\% for No-Memory, and 8.06\% for Only-Thought. For the final 20 updates, we also calculate the mean cosine similarity between the text vector of each generated thought and that of the initial thought; the corresponding values are .9512, .9133, and .9322. For both measures, the comparisons between Full and No-Memory have two-sided Holm-adjusted $p=2.38\times10^{-7}$. However, the proportions of successive thoughts whose text-vector cosine similarity is at least .995 are 57.92\% for Full, 25.38\% for No-Memory, and 49.72\% for Only-Thought.

Complete protocols and additional results, including those for Heuristic-Only, are reported in Appendix~\ref{app:Controlled_Empirical_Validation}.

\section{Conclusion}

We presented IFRGA to generate facial reactions by maintaining and updating a latent state. Specifically, LSEM uses a mechanism inspired by human thought processes to revise IFRGA's interpretation using memory, retained thoughts, and current speaker behaviour. FRGM then generates facial reactions using the current thought and emotion together with speaker behaviour. On REACT 2025, IFRGA achieves a higher FRCorr and a lower FRDist than the published results of the other learnt methods in our comparison. Furthermore, in a blinded human evaluation, facial reactions generated while carrying thoughts across matched updates received higher ratings for naturalness and contextual appropriateness than those generated while retaining factual summaries. 
In future work, we will explore more efficient latent state updates for longer interactions. We will also investigate how to adapt IFRGA to different interaction contexts and virtual avatars.

\section*{LLM Usage Statement}

The authors used GPT-6 Astra and Gemini 4 Argon strictly for assisting with code and English language polishing, grammar correction, and improving readability during the preparation of this manuscript. All conceptual ideas, methodology formulations, mathematical designs, experimental implementations, and data analyses were independently developed and conducted by the human authors. Large language models (such as Qwen3-8B and DeepSeek) mentioned in the methodology and evaluation sections were utilized strictly as experimental components and evaluators as described in the paper. All text and citations were thoroughly verified, edited, and validated by the authors.

{
    \small
    \bibliographystyle{ieeenat_fullname}
    \bibliography{sn-bibliography}
}

\appendix
\section{The Psychological and Philosophical Foundations of Intentional Agents}\label{app:philosophical_foundation}

The Facial Reaction Generation (FRG) task has made important progress in cross-modal alignment and visual fidelity. However, the vast majority of existing works adopt an end-to-end deep learning paradigm. From the perspective of human psychology and cognitive science, this paradigm essentially falls into the reductionist trap of Behaviourism: it downgrades the complex facial reactions of a listener into a pure Stimulus-Response mechanism, attempting to establish a direct black-box mapping between the speaker's audio-visual features and the listener's facial muscle movements.

In authentic non-verbal social interactions, human facial expressions are never merely passive reactions to external signals. End-to-end models attempt to forcefully bypass the most core component of interaction: the internal cognitive appraisal process \citep{scherer2001appraisal,scherer2005emotions}. Consequently, while the generated expressions may achieve frame-level visual realism, they lack coherent logical depth in long-horizon social contexts. In fact, the visible tip of the iceberg accounts for less than one percent of its entirety. The facial reactions exhibited by humans are fundamentally the subconscious projection of complex latent state processes. Therefore, to achieve meaningful and interpretable facial reaction generation akin to human behaviour, it is imperative to shift the focus toward simulating latent state dynamics.

``Intentionality'' is a philosophical concept. In traditional Husserlian phenomenology, Intentionality is often statically defined as ``consciousness of something,'' emphasising the subject's passive perception and representation of external objects \citep{husserl2012ideas}. However, in the context of this paper, intentionality implies first constructing a self-sustaining ``Internal Stream of Consciousness.'' Within this continuously flowing latent state, the agent spontaneously generates a directedness toward the external world, which ultimately translates into embodied action outputs. This aligns with the concept of ``Operative Intentionality'' proposed by Merleau-Ponty in embodied phenomenology \citep{merleau2013phenomenology}. Operative intentionality is not abstract contemplation enclosed within the brain, but rather a pre-reflective vital force, directly driven by internal bodily tension and extending into the world. This differs from the concept of ``proactivity'' frequently discussed in current AI agent literature. These so-called ``proactive'' capabilities remain essentially reactive. In practice, conventional proactivity merely means the agent exhibits the capacity for spontaneous triggering and action planning driven by external inputs and environmental cues. Its root cause still lies in exogenous algorithmic mechanisms and interactive feedback structures.

\begin{figure*}
  \centering
  \includegraphics[width=\linewidth]{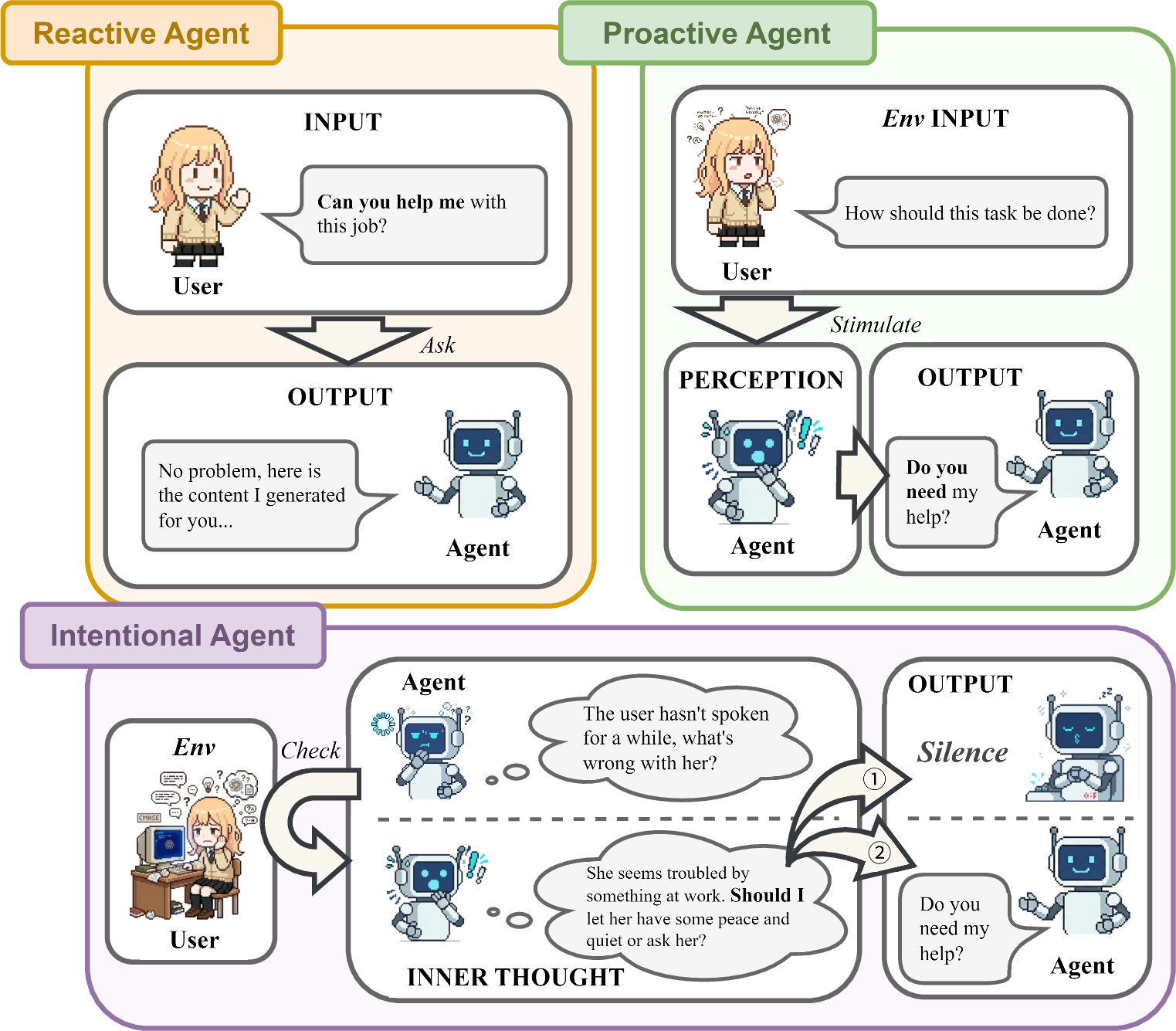}
  \caption{\textbf{Comparison of agent systems across three paradigms of human-agent interaction.} Reactive agents passively receive user queries and then generate responses. Proactive agents infer tasks based on environmental observations and propose possible assistance requests accordingly. Intentional agents maintain a continuous latent state trajectory driven by endogenous intentionality, allowing them to evaluate complex contexts (such as user silence) and autonomously decide whether to initiate action or remain silently active.}
  \label{fig:agent_paradigms}
\end{figure*}

Figure \ref{fig:agent_paradigms} illustrates the comparison among intentional, proactive, and reactive agents. Proactivity frees agents from relying on passive inputs. In contrast, intentionality requires agents to possess an autonomous generative mechanism within their latent state dynamics.

In other words, current proactivity only solves the problem of ``how to act'' under specific conditions, while lacking an endogenous drive when it comes to ``why to act.'' This limitation becomes particularly glaring in real-time, natural conversations, especially in scenarios involving extensive non-verbal intervals and social silence. When real-time interaction enters a blank period lacking explicit external commands or environmental cues, this exogenously driven logic often leads to a breakdown in cognitive coherence and a stagnation in behavioural generation. This prompts us to rethink the architectural bottleneck: rather than merely simulating proactivity, we should introduce an endogenous operative intentionality in the architecture. That is, the system should be able to internally generate an understanding of objects, goals, and values, and utilise the body to extend this understanding outward.

Naturally, we must first define the body of the agent. Lan \cite{lan2018general}, in his research on digital capitalism, proposed the concept of the ``virtual-body''. According to Lan, just as we possess a physical ``natural body'' in reality, we also inhabit a ``virtual-body'' when engaging with internet platforms and digital spaces. This virtual-body is composed of computable, identifiable, and trackable data. For instance, to a social media user, the account is merely an intermediary for participation. However, to the platform itself, the account is the only truly existent object. The platform populates this initially ``empty formal set'' with content (clicking data, emotional expressions, and attentional investment), thereby ``fleshing out'' the digital presence. Between the user's choices and the platform's corresponding shaping, the presence of this virtual-body becomes increasingly prominent.

For agents, the ``virtual-body'' theory similarly reveals their inherent corporeality. However, a crucial distinction must be made here between the ``body'' and the ``soma''. In natural language, ``body'' often refers to the physical manifestation, that is, the soma. This is the material substrate of a living entity. The soma and its physiological processes can only be perceived from an external, third-person perspective with a naturalistic attitude. Yet, human vital expression is simultaneously internal and external, encompassing both experience and behaviour. Therefore, from a phenomenological perspective, there must exist a ``body'' that is inextricably linked to first-person experience, that is, the flip side of the soma. This ``body'' is connected to the lifeworld through perception and is by no means confined within the material limits of the soma. This distinction reminds us that consciousness is not encapsulated within an organism but enters the world through the openness of the body.

For a digital agent, its underlying neural network infrastructure, data structures, and the mesh/rigging of its 3D avatar constitute merely its ``soma''. The soma is the material substrate of the system and must be observed from a third-person naturalistic perspective. However, when these somatic substrates are endowed with an internal circulation mechanism within the digital interaction network, the agent truly acquires a body for perception, communication, and interaction, that is, the virtual-body. By establishing the agent’s virtual-body, we open a specific field for the operation of intentionality. It transforms the agent from a dead object statically awaiting inputs into an entity maintaining continuous latent state dynamic tension. In our architecture, this tension is instantiated as the continuously self-updating Inner Thought Flow and multi-dimensional emotion drives.

Within this framework, our understanding of Emotion undergoes a fundamental reconstruction. As pointed out by phenomenological psychiatrist Thomas Fuchs \cite{fuchs2017ecology}, the essence of emotion is ``e-motion'', where the prefix ``e-'' emphasises ``ex-'' or ``outward'', that is, an outward movement. This implies that facial expressions are by no means mechanical reactions to external environmental stimuli, but rather action outputs where the internal ``stream of consciousness,'' having crossed a certain homeostatic threshold, spontaneously extends toward the external world seeking a physical outlet.

Consequently, human-like facial reaction generation can be redefined as an ``inside-out'', continuous, and dynamic emergent process. External interaction contexts (such as the speaker's multimodal inputs) merely act as perturbation variables flowing into the agent's internal stream of consciousness. Even when external inputs are entirely null (e.g., prolonged social silence), driven by the psychological inertia and internal cycles induced by operative intentionality, the agent continues to undergo state transitions, persistently projecting this internal stream of consciousness (e-motion) into visible facial muscle movements.

In summary, the intentionality introduced in this paper can be viewed as the coupled outcome of a three-tier mechanism: the soma provides potential and constraints; the virtual-body opens up a dwelling place within the lifeworld; and the latent state dynamics cycle and affective mechanisms endow actions with directedness. Thus, this intentionality is not a static property, but an emergent dynamical process. It enables the agent to spontaneously generate behaviour in the absence of external stimuli. Its facial movements are no longer passive reflections of external audio-visual signals, but the spontaneous physical extension (e-motion) of internal cognitive tension.

\section{Theoretical Basis and Validation of Latent State Evolution}
\label{app:eq_validation}

\begin{figure}[t]
  \centering
  \includegraphics[width=\columnwidth]{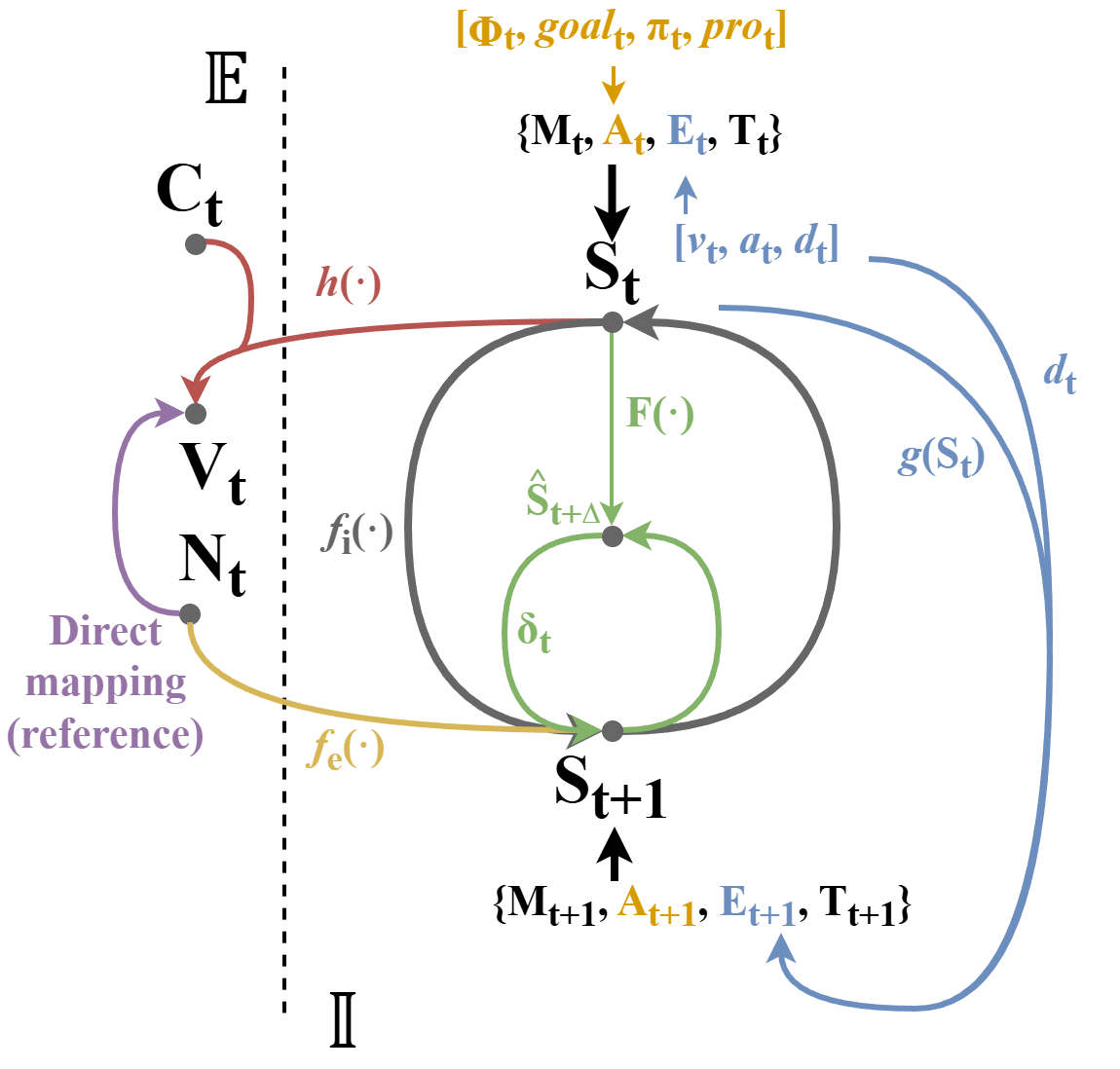}
  \caption{\textbf{Latent state evolution.} The diagram provides a conceptual illustration of latent state updates and facial reaction generation. Facial reaction generation also uses speaker behaviour, as described in the main paper.}
  \label{fig:internal_dynamics}
\end{figure}

Figure~\ref{fig:internal_dynamics} illustrates the conceptual latent state update described in this section.

\subsection{Heuristic Status and Computational Requirements}
The following expressions provide a heuristic description informed by appraisal theory, predictive processing, and the VAD model. They are not uniquely derived from these theories or from a single probabilistic model or objective. They describe four computational requirements:

\begin{enumerate}
    \item \textbf{Autonomous continuation}: Latent states must be capable of continuing their temporal evolution in the absence of novel external stimuli ($N_{t+1} = \varnothing$).
    \item \textbf{Action-expectation binding}: Action formulations must incorporate both an immediate behavioural choice and its anticipated perceptual outcome.
    \item \textbf{Error-driven correction}: Discrepancies between anticipations and realisations must generate scalar prediction errors that correct subsequent latent states.
    \item \textbf{Embodied physical realisation}: Abstract latent state affect must be projected outward under concrete physical and avatar rig constraints.
\end{enumerate}
Guided by these operational requirements, we chose compact, interpretable, and testable functional forms rather than claiming a unique mathematical realisation of Appendix~\ref{app:philosophical_foundation}. In this appendix, $N_{t+1}$ denotes the observation used to update $S_t$; the main paper denotes this observation by $N_t$. The no-change marker represents the absence of new information.

\subsection{Equation-wise Theoretical Basis and Design Rationale}
Table~\ref{tab:eq_basis_rationale} summarises the theoretical basis and purpose of each step. The descriptions below cover the state update and facial generation.

\begin{table*}[t]
\centering
\caption{Theoretical basis and design rationale for state evolution and facial generation.}
\label{tab:eq_basis_rationale}
\small
\begin{tabularx}{\linewidth}{l p{5.2cm} X}
\toprule
\textbf{Step} & \textbf{Theoretical Basis} & \textbf{Functional Design Rationale} \\
\midrule
Candidate state & Self-sustaining dynamics in Appendix \ref{app:philosophical_foundation}
& Separates internal and input-driven changes, making silence updates testable. \\

Action state & ITCMA \citep{zhang2025itcma}
& Separates action choice, purpose, feasibility, and expected result. \\

Prediction cycle & Predictive coding \citep{rao1999predictive}
& Gives an explicit prediction, comparison, and correction cycle. \\

Emotion update & Appraisal dynamics and EMA \citep{marsella2009ema}
& Retains emotional continuity while allowing appraisal and error to change emotion. \\

Facial projection & VAD \citep{russell1977evidence,mehrabian1996pleasure} and embodied intentionality & Separates affective meaning from character-specific facial realisation. \\
\bottomrule
\end{tabularx}
\end{table*}

\paragraph{Candidate State Generation.}
The candidate update $\tilde{S}_{t+1} = f_i(S_t) + f_e(S_t, N_{t+1})$ decomposes state evolution into latent state persistence $f_i(S_t)$ and perceptual modulation $f_e(S_t, N_{t+1})$. Crucially, the addition operator conceptually separates these two sources rather than indicating element-wise addition over heterogeneous text and vector representations. When external inputs cease ($N_{t+1} = \varnothing$), external modulation vanishes, $f_e(S_t, \varnothing) = 0$, while $f_i(S_t)$ can continue the state update. This formalisation motivates the controlled no-change probes. The state-channel ablation examines how retained inputs affect subsequent updates under its specified execution feedback. External perception is supplied as $N_{t+1}$, while prediction error $\delta_t$ is routed to subsequent state correction; both inform the Inner Thought Flow (ITF) without persisting as text fields in $S_t$.

\paragraph{Action State Factorisation.}
The action state $A_t = \{\pi_t, \mathrm{goal}_t, \phi_t, \mathrm{pro}_t\}$ gives a conceptual grouping of action tendency, purpose, feasibility, and expected outcome, inspired by ITCMA~\citep{zhang2025itcma}. The runtime schema exposes the selected action text and a semantic expectation, together with the next thought and a reported scalar prediction error. Purpose and feasibility are expressed in the thought and action descriptions. These records support inspection of the proposed action and its anticipated consequence.

\paragraph{Predictive Coding Cycle.}
The following expressions follow the prediction, error, and correction order of predictive coding:
\begin{enumerate}
    \item \textbf{Prediction}: $\hat{S}_{t+\Delta}=F(S_t)$ produces the expected part of a future state rather than predicting every field. In the implementation, the base large language model produces a semantic expectation for comparison with a later observation.
    \item \textbf{Comparison and Error Generation}:
    $\delta_t=C_{\mathrm{LLM}}(\hat{S}_{t+\Delta},N_{t+\Delta})$ represents the implemented comparison between the previous semantic expectation and the later observation. The base large language model performs this comparison and returns a scalar prediction error $\delta_t$. This avoids treating heterogeneous latent states as vectors that can be directly subtracted. Following predictive coding, a stronger conflict produces a higher error.
    \item \textbf{Error-Conditioned State Propagation}:
    $S_{t+1}\leftarrow J(\tilde{S}_{t+1},\delta_t)$ summarises error-dependent state propagation. The runtime retains the new error returned jointly with thought, action, and expectation. This retained error enters the following programmed affect update. Memory preparation and numerical affect updating precede each language-model call.
\end{enumerate}

\paragraph{Emotion Update.}
The emotion update follows appraisal dynamics by treating emotion as a continuous state while making the implemented prediction-error input explicit \citep{marsella2009ema}:
\begin{equation*}
\begin{aligned}
E_{t+1} &= \operatorname{clip}\left(\alpha(\delta_t)E_t+[1-\alpha(\delta_t)]g(S_t,N_{t+1},\delta_t),\,-1,\,1\right),\\
\alpha(\delta_t) &= 0.7(1-0.5\delta_t).
\end{aligned}
\end{equation*}
The programmed appraisal operator combines text-derived VAD values with the retained affect. Let $\mathcal{V}$ denote the configured text-to-VAD estimator, and let $x^{(P)}$, $x^{(M)}$, and $x^{(T)}$ denote the appraised perception, memory or current-conversation history, and retained thought. The implemented appraisal has the form
\[
g(S_t,N_{t+1},\delta_t)
=\sum_{i\in\{P,M,T\}}w_i\mathcal{V}(x^{(i)})
+(0,\,0.4\delta_t,\,-0.5\delta_t)^{\mathsf T}.
\]
The weights are $(.6,.15,.25)$ for populated perception text and $(0,.4,.6)$ when the appraised perception is empty or a literal \texttt{No change} or \texttt{None} marker. This branch uses the actual text supplied to the estimator. The retained main configuration uses the NRC analyser. Contextual variants use the frozen EmoBank regressor and the appraisal-text scope recorded by their protocols. The coefficients .7 and .5 in the inertia function, and .4 and .5 in the error adjustment, are empirical choices. The new error reported by the current LLM call enters the following affect update. The initial error is set to zero when a previous expectation is unavailable.

\paragraph{Facial Projection.}
Facial generation uses the latent state and speaker behaviour under character-specific constraints. A learnt speaker-conditioned low-frequency mapping supplies base facial values, and a Diffusion Transformer supplies learnt residual dynamics. A character-specific adaptor converts the resulting Action Units into avatar BlendShapes.

\subsection{Overall Functional Pipeline}
The equations specify the information sources and functional dependencies of the update. The runtime first prepares memory and updates the affective state using the prediction error reported in the preceding update. One successful LLM call then returns the next thought, selected action, semantic expectation, and current prediction error. The generated fields and updated affect are retained for the following update and supplied to the configured facial readout. Prediction and comparison therefore share the state-generation call, while memory preparation and numerical affect updates are programmed operations.

\subsection{Controlled Empirical Validation on Component Structures}\label{app:Controlled_Empirical_Validation}
We evaluate these functional designs through controlled state-role tests, facial and state ablations, and long-horizon tests of operational stability and free-thought semantic continuity.

\paragraph{1. State Roles and Directed Interventions.}
We compare explicit state-role labelling using eight constructed contexts, 16 factorial state combinations, three language seeds, and three input formats, yielding 1152 model calls. Full uses separate labelled fields. Labelled-Flat contains exactly the same field values, role labels, rules, and options in one paragraph. Unlabelled-Flat removes the role labels. All formats use a 512-token output cap.

\begin{table*}[htbp]
\centering
\caption{State-role controls on constrained action selection and expectation comparison.}
\begin{tabular}{lrrrr}
\toprule
Format & Balanced accuracy & Required-action success & Incorrect action when waiting & Conflict Brier \\
\midrule
Full & .98958 & .98958 & .01042 & .00000 \\
Labelled-Flat & .98611 & .97917 & .00694 & .00000 \\
Unlabelled-Flat & .94792 & .95833 & .06250 & .11719 \\
\bottomrule
\end{tabular}
\end{table*}

\paragraph{2. Expectation Comparison and Intervention Responses.}
The three predefined comparisons use contexts as paired units and apply Holm correction. Relative to the mean of the two labelled formats, removing role labels increases conflict Brier by .11719, with a 95\% interval of [.07292, .17188] and $p_{\mathrm{Holm}}=.02344$. Full versus Labelled-Flat gives adjusted $p=1.000$ for both balanced action accuracy and conflict Brier.

For single-factor interventions on permission, ongoing execution, current goal, and expected conflict, Full produces correct actions for both members of 97.92\% of pairs. The absolute deviation of the predicted conflict-error change is zero for each intervention factor. These results support explicit role information and directed state interventions in the specified controlled task.

\paragraph{3. Facial Generation Ablation and Low-Frequency Guidance.}
The current Full condition combines a learnt speaker-conditioned low-frequency mapping with a neural residual. Learnt-Only is the independently trained control without the mapping, while Heuristic-Only retains the original hand-crafted rules. We evaluate 60 contexts with three training seeds, averaging seeds within each context. Low-frequency trajectories use a Gaussian filter with $\sigma=5$ frames. Phase boundaries follow the actual audio clock, yielding 57 valid long-silence contexts.

\begin{table}[t]
\centering
\small
\setlength{\tabcolsep}{3pt}
\caption{Low-frequency AU-MAE. Paired tests compare Full with Learnt-Only.}
\label{tab:low-frequency-au-mae}
\begin{tabular}{@{}lrrrr@{}}
\toprule
Phase & Full & \shortstack{Learnt-\\Only} &
\shortstack{Heuristic-\\Only} & $p$ \\
\midrule
All & .23403 & .25143 & .22868 & .01812 \\
Long silence & .22728 & .27599 & .26313 & .00905 \\
\bottomrule
\end{tabular}
\end{table}

Full improves low-frequency error in 39 of 60 contexts overall and 36 of 57 long-silence contexts. Its full-sequence high-frequency AU-MAE is .05775, compared with .08876 for Learnt-Only ($p=1.63\times10^{-11}$). The mean cosine similarity between the change from Learnt-Only to Full and the original hand-crafted guide is .04095, with positive values in 40 of 60 contexts ($p=.00133$). The current trajectory results support an average contribution of the low-frequency branch to slow facial trends.

\paragraph{4. State Ablation during Speaker Silence.}
We start from 30 predefined long-silence candidates. Six contexts are used for development. Applying the same input-based requirement of at least three seconds of actual speaker silence to the remaining 24 candidates yields 16 confirmation contexts.

Each context uses three language seeds. All six state conditions start from the same Full prefix state and perform real model updates at 1, 2, and 3s after silence onset. Time advances logically. The experiment contains 48 shared-prefix calls and 864 subsequent calls, yielding 912 actual calls. Language seeds are averaged within each context before paired analysis over the 16 original recordings.

The task uses the listener role from the recorded conversation. At silence onset, a simulated executor receives the initial Action. Execution lasts $\max(2,w/3)$ seconds, where $w$ is the command's word count. Conditions that retain perception receive feedback on whether execution is ongoing or complete; a new command replaces an unfinished command. A non-None Action initiates a new command, while \texttt{None} continues the current activity.

Full retains perception, memory, emotion, thought, and expectation. Only-X retains the specified direct input slot. The original episodic memory implementation is used, and a retained memory slot includes its historical thought, action, and expectation content. Emotion uses frozen contextual VAD estimates and the original state-update equations. The initial prediction-error input is set to zero, with the first raw model report stored separately.

\begin{table*}[htbp]
\centering
\caption{State-channel ablation during the first three seconds of speaker silence. Prediction error is the model's self-reported scalar.}
\begin{tabular}{lrrr}
\toprule
Condition & New-command rate & Self-reported prediction error & Simulated interruptions per three updates \\
\midrule
Full & .000000 & .055556 & .000000 \\
Only-Perception & .000000 & .247917 & .000000 \\
Only-Memory & .041667 & .193750 & .041667 \\
Only-Emotion & .000000 & .137847 & .000000 \\
Only-Thought & .125000 & .450694 & .104167 \\
Only-Expectation & .027778 & .572222 & .020833 \\
\bottomrule
\end{tabular}
\end{table*}

Seven primary comparisons jointly receive Holm correction. Four compare command rates against Only-Perception, Only-Emotion, Only-Thought, and Only-Expectation. Three compare self-reported prediction error against Only-Emotion, Only-Thought, and Only-Expectation.

\begin{table}[htbp]
\centering
\small
\caption{Holm-adjusted primary comparisons against Full.}
\begin{tabular}{llr}
\toprule
Outcome & Control & Holm $p$ \\
\midrule
Command rate & Only-Perception & 1.000000 \\
Command rate & Only-Emotion & 1.000000 \\
Command rate & Only-Thought & .012624 \\
Command rate & Only-Expectation & .539137 \\
Prediction error & Only-Emotion & .038707 \\
Prediction error & Only-Thought & .003920 \\
Prediction error & Only-Expectation & .003037 \\
\bottomrule
\end{tabular}
\end{table}

Full has zero new-command rate, tied with Only-Perception and Only-Emotion. Its command rate is significantly lower than Only-Thought, and its mean self-reported prediction error is the lowest among the six conditions. The three specified error comparisons survive correction. Simulated interruptions are reported descriptively.

These results support constrained action initiation and lower self-reported surprise within the first three seconds of silence under the specified execution-feedback assumptions.

\paragraph{5. Long-Horizon Silence Stability.}
After initialisation, we stop introducing new perceptual content and run 120 real state updates over a 30-minute logical horizon. The measured wall-clock duration is 1983.52s. All updates produce structurally valid states and finite facial outputs. Rule-based checks record zero unsupported factual external events.

The monitored process reaches a peak resident set of 2.178GB, with an end-to-start change of $-86.942$MB. Mean similarity between adjacent thoughts is .99268. These measurements establish structural, numerical, and process-memory stability over the tested run.

\paragraph{6. Semantic Continuity during Long Silence.}
We construct 24 new everyday contexts, comprising eight positive, eight negative, and eight uncertain situations. Each context contains four predefined groups of factual word cues. Full, No-Memory, and Only-Thought receive identical initial thoughts containing all initial facts. A common prompt encouraging concrete current reasoning is selected using two earlier development contexts.

Full retains memory, emotion, expectation, and thought. No-Memory retains emotion, expectation, and thought while removing persistent memory. Its affective appraisal can use the permitted preceding thought. Only-Thought carries the preceding thought. During the experiment, external perception supplies no new observations, and the agent freely updates thought.

Each context uses three language seeds and 80 updates per condition, yielding 17,280 actual model updates. Each update advances logical time by 22.5s, producing a 30-minute logical horizon. Seeds are averaged within each context before paired analysis.

The two predefined primary endpoints are mean coverage of the four fact-term groups and context cosine similarity over the final 20 updates. Fact-term coverage measures the fraction of predefined cue groups mentioned in each thought. Context similarity is the cosine similarity between the text vectors of the generated thought and the initial thought, computed using spaCy.

\begin{table*}[htbp]
\centering
\small
\caption{Free-thought continuity with identical complete initial facts.}
\begin{tabular}{lrrr}
\toprule
Measure & Full & No-Memory & Only-Thought \\
\midrule
Fact-term coverage & .20885 & .03077 & .08064 \\
Final-20 context cosine & .95118 & .91332 & .93225 \\
Any anchor term present & .83958 & .16684 & .38299 \\
Near-exact repetition & .57917 & .25382 & .49722 \\
Rule contradiction rate & .00000 & .00000 & .00000 \\
Rule unsupported-event rate & .00000 & .00000 & .00000 \\
\bottomrule
\end{tabular}
\end{table*}

Full exceeds No-Memory on fact-term coverage in 23 of 24 contexts and on final context similarity in all 24 contexts. Both primary two-sided comparisons have Holm-adjusted $p=2.38\times10^{-7}$. Supplemental Full versus Only-Thought comparisons give adjusted $p=8.34\times10^{-6}$ for fact-term coverage and $2.38\times10^{-7}$ for final context similarity.

Full has a near-exact repetition rate of 57.92\%, compared with 25.38\% for No-Memory and 49.72\% for Only-Thought. All conditions have zero contradiction and unsupported-event rates under the predefined word-based checks. These results support persistent memory for maintaining the measured factual cues and context continuity during free thought.

\subsection{Matched-Update ITF Control}
\label{app:matched_itf_control}

\paragraph{Original human-study update and input control.}
We use all 24 recording contexts from the existing random selection and two language-model seeds per context. Both conditions use Qwen3.5-9B, identical system and user templates, complete observed history, a shared initial state, and logical updates at 0, 2, 4, and 6 seconds within an eight-second window. Each update has a 256-token output cap and an 8192-token context cap. The initial call is shared, followed by three calls per condition. These logical times specify the offline component test.

Full carries its preceding generated thought. The control replaces this carried content with an observation-based factual summary. Both conditions generate a fresh current-step thought and continue to update affect, expectation, and action. The factual summary is stable when observations provide no new facts. Both conditions use the same facial checkpoint and paired diffusion seeds. For human presentation, language-model seeds alternate by the pre-existing context index. All 24 contexts are retained, including 14 pairs with different final AU trajectories and 10 with identical trajectories.

\paragraph{Controlled zero-input definition.}
After the initial context, new utterances, audio features, video frames, and visual descriptions are stopped. Subsequent updates receive fixed no-change markers and retained state. Natural speech gaps with continuing visual input are separately described as speech silence.

\paragraph{Matched-update intention retention.}
We use 14 real contexts whose original recordings are disjoint from training and validation. Each context contains two controlled intention pairs, concerning requesting further detail versus reflecting existing content, and providing practical advice versus first providing emotional support.

Full, Factual-Summary, and Goal-Summary share the initial thought, memory, emotion, and expectation for each initial intention. Full carries its generated preceding thought. Factual-Summary carries observed facts. Goal-Summary carries observed facts and the fixed current intention.

Each condition uses two language seeds, logical times 0, 2, 4, and 6s, three actual updates, and a shared 180-token output cap, yielding 1008 model calls. For each terminal thought, we subtract its cosine similarity to the alternative initial intention from its cosine similarity to the selected intention. Direction accuracy is the proportion of terminal thoughts for which this difference is positive. Intention pairs and seeds are averaged within each recording before analysis.

\begin{table}[htbp]
\centering
\caption{Intention retention under matched updates and output budgets.}
\begin{tabular}{lrr}
\toprule
Condition & Cosine difference & Direction accuracy \\
\midrule
Full & .005902 & .66071 \\
Factual-Summary & -.000662 & .50000 \\
Goal-Summary & .015320 & .75000 \\
\bottomrule
\end{tabular}
\end{table}

Full exceeds Factual-Summary by .006563 in this cosine difference, with a 95\% recording-bootstrap interval of [.004595, .008684] and Holm-adjusted $p=.000488$. Goal-Summary exceeds Full by .009419, also with adjusted $p=.000488$.

Goal-Summary reconstructs its carried input at every update by appending the initial intention verbatim to the observed facts. This keeps the reference intention explicitly available throughout the experiment. Full instead carries its newly generated preceding thought. The measured endpoint assesses similarity to the initial intention relative to an alternative intention, making persistent goal availability directly relevant to the score.

All conditions share update times, a 180-token output cap, and an 8192-token context limit. Mean recorded input-token counts are 872.37 for Full, 957.60 for Factual-Summary, and 976.77 for Goal-Summary. Input length and carried information therefore differ across conditions.

These results support an advantage of thought carry over fact-only carry for retaining the selected intention. Explicit goal reintroduction provides stronger retention on this endpoint. The separate human comparison evaluates Full against factual-summary carry using the same model configuration.

\paragraph{Terminal-state facial readout.}
We evaluate the terminal states from this experiment using a fixed eight-second facial condition. Speaker attributes, audio, facial weights, and paired random numbers are shared between the two initial intentions. Only the terminal state supplies the state-dependent facial condition.

Each of the 336 terminal states generates ten distinct trajectories, yielding 3360 actual predictions. Mean absolute AU differences between the two initial intentions are .001724 for Full, .000276 for Factual-Summary, and .001077 for Goal-Summary.

The Full-minus-Factual-Summary difference is .001448, with a 95\% interval of [.001099, .001900]. The Full-minus-Goal-Summary difference is .000647. Both recording-level comparisons have Holm-adjusted $p=.000244$. These results support numerical modulation of terminal facial parameters by initial intention carried through subsequent state updates.

\section{Details of Facial Expression Mapping}\label{app:facial_mapping}

\begin{figure*}[htbp]
  \centering
  \includegraphics[width=\linewidth]{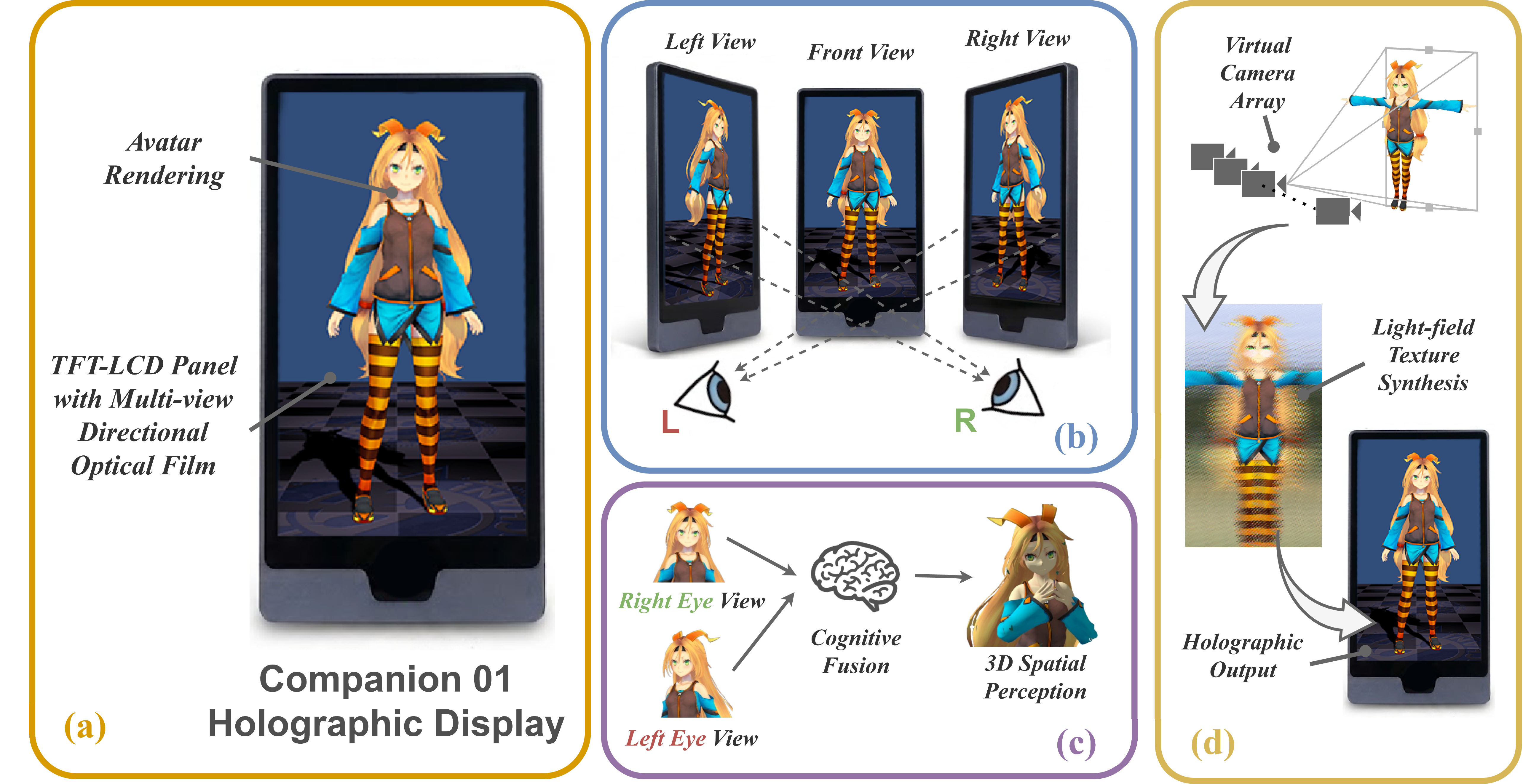}
  \caption{\textbf{Holographic Avatar display.} (a) Physical presentation of the Companion 01 holographic display; (b) The multi-view directional optical film on the TFT-LCD panel directs images of different perspectives; (c) Utilising the binocular parallax principle, the left and right eyes receive different views which are cognitively fused in the brain to achieve a naked-eye 3D spatial perception; (d) The rendering pipeline, where a virtual camera array captures multi-view images that are processed via light-field texture synthesis to generate the final holographic output.}
  \label{fig:holographic_avatar}
\end{figure*}

To enable the proposed agent framework to support real-world human-computer interaction, this study constructed a display and behaviour control system for the avatar based on the Unity 3D engine. At the system architecture level, an asynchronous REST API server is embedded within the frontend presentation layer. This serves as a communication hub between the underlying engine and the upper-level agent framework to receive and execute action and state manipulation commands issued by the agent hub. On the physical presentation side, the system adopts the Companion 01 device as the holographic display carrier. This device integrates a TFT-LCD panel and is covered with a multi-viewpoint directional optical film on its surface. It utilises the principle of binocular parallax to achieve a glasses-free holographic visual effect as shown in Figure \ref{fig:holographic_avatar}.

The system constructs a set of horizontally arranged microlens virtual camera arrays within the rendering pipeline. It aligns the optical axes of the cameras to intersect in real time at the core position of the avatar to lock the focal plane. During each frame rendering cycle, the camera array synchronously captures the scene and generates a sequence of images with slight perspective differences. Subsequently, the swizzle algorithm accurately extracts corresponding pixels from the multi-viewpoint images for spatial reorganisation based on the physical refraction characteristics of the optical film on the screen surface. Finally, it synthesises a composite texture containing the full light field information and outputs it to the physical screen. To enhance the embodied interaction sense of the virtual avatar, the display system integrates a visual perception module based on the frame difference method to capture salient motion areas in the real environment. This system can not only guide the visual attention of the virtual avatar in real time but also parameterise external environmental stimuli and feed them back to the agent framework. This subsequently influences the thinking logic and emotional state of the agent.

The current facial generator uses thought semantics, retained affect, and speaker behaviour. A learnt speaker-conditioned low-frequency mapping and a trained DiT residual jointly produce facial parameters. The numerical readout is specified in Appendix~\ref{app:emotion_mapping}.

The Unity display prototype provides a character-specific rig and can apply breathing and body-sway overlays. Display settings are configured separately from the numerical AU generator. In the additional human studies, all compared clips use a common browser avatar, camera, and AU amplitude, with additional idle animation disabled. The studies therefore compare the supplied AU trajectories under common presentation conditions. Their protocol is reported in Appendix~\ref{app:additional_human_protocol}.

\section{Details of Emotion Vector Mapping}\label{app:emotion_mapping}

The current facial readout uses thought semantics and V/A values from retained state together with speaker behaviour. This section reports text-to-VAD validation, the main state-to-facial readout, and affect-estimator replacement in a separately trained Action-conditioned facial component.

\subsection{Validation of Text-to-VAD Mapping}
\label{app:text_vad_validation}

The retained main benchmark configuration obtains text-derived appraisals through the NRC VAD analyser~\citep{mohammad-2018-obtaining,mohammad2025nrc}. Its agreement with human sentence-level affect is evaluated on EmoBank~\citep{buechel-hahn-2017-emobank}. The controlled contextual variants use a frozen text encoder and an EmoBank-trained regressor, with their configuration and validation reported in Appendix~\ref{app:vad_replacement}. The VAD state is maintained by the programmed affect update. Speaker facial V/A values provide observations about the speaker, and the retained VAD state describes the agent's own appraisal.

Human VAD scores in EmoBank range from 1 to 5 and are linearly converted to $[-1,1]$. Evaluation uses the official 1000-sentence test set. Direction accuracy uses clearly non-neutral sentences, with 489, 456, and 382 eligible sentences for Valence, Arousal, and Dominance, respectively.

\begin{table*}[htbp]
\centering
\caption{Validation of the NRC text-to-VAD analyser against human EmoBank annotations.}
\label{tab:text_vad_validation}
\begin{tabular}{lrrrrr}
\toprule
Dimension & Pearson & Spearman & Direction accuracy & Eligible sentences & Binomial $p$ \\
\midrule
Valence & .5119 & .4835 & 67.28\% & 489 & $8.34\times10^{-15}$ \\
Arousal & .2358 & .2230 & 58.77\% & 456 & .000104 \\
Dominance & .1507 & .1473 & 66.23\% & 382 & $1.09\times10^{-10}$ \\
\bottomrule
\end{tabular}
\end{table*}

All three dimensions show significant Pearson and Spearman correlations with human annotations and above-chance direction accuracy. Valence has the strongest agreement. These results support text-derived VAD as a low-dimensional affective condition.

\subsection{Evaluated State-to-Facial Readout}

The retained main benchmark configuration uses the current thought and the agent's retained V/A values. Let $\mathbf u_f\in\mathbb R^{300}$ denote the semantic condition at frame $f$. It uses the current thought embedding. When an observed speaker word is active, its embedding is averaged with the thought embedding over the aligned word interval. The retained V/A values are interpolated on the facial time axis.

The learnt low-frequency model uses 72 features. These comprise 25 smoothed speaker facial attributes, $(v_f,a_f,v_f^2,a_f^2,v_fa_f)$ from the agent's affect, ten speaker acoustic features, and 32 projected speaker-transcript features. It predicts a 25-channel base $\hat{\mathbf r}_f$, comprising 15 AU values, two facial V/A values, and eight expression probabilities. The DiT condition is
\[
\mathbf c_f=
[\mathbf u_f,\,3v_f,\,2a_f,\,\hat{\mathbf r}_{f,1:15}]
\in\mathbb R^{317}.
\]

The retained facial checkpoint was trained on 1333 official training clips, with 327 internal training clips used for checkpoint selection. Its cached semantic and V/A conditions were derived from paired speaker and listener transcripts, with listener speech taking precedence in overlapping intervals. The physical training condition was the first 15 channels of the true training reaction's Gaussian low-frequency component. Its 25-channel target was the remaining high-frequency component. Training used noise-prediction loss and a reconstructed-sample CCC term. These training conditions and targets are recorded with the checkpoint.

At inference, the semantic and affect channels receive the mixture of thought and speaker-word embeddings and retained V/A, while the low-frequency AU condition comes from the predicted base. The DiT generates a 25-channel residual using 20 DDIM steps and CFG=3. A high-pass projection removes slow residual drift, and per-channel training scales limit its standard deviation. The main configuration adds all 25 projected residual channels to the base and applies Gaussian smoothing.

The retained readout also contains a training-derived residual mean and a stochastic AU-probit transformation. Let $\mathbf b_f$ denote the smoothed sum of the base and residual including that mean, and let $\Phi$ be the standard normal cumulative distribution function. For an AU channel $j$ with fitted latent variance above $10^{-8}$, the readout is
\[
p_{fj}=\operatorname{clip}_{[0,1]}(b_{fj}),\qquad
a_{fj}=\Phi\!\left(
\frac{\Phi^{-1}(p_{fj})+\rho z_{fj}}{\sqrt{1-\rho^2}}
\right),\qquad \rho=.55.
\]
The latent $z_{fj}$ is normalised using the sum of fitted style and temporally correlated motion variances for channel $j$. The CDF transformation uses its limiting values at probabilities zero and one. Channels with fitted latent variance at most $10^{-8}$ retain $p_{fj}$ directly.

Independent facial predictions sample both diffusion initial values and this latent process. The additional additive motion coefficient is zero. Facial V/A values are constrained to $[-1,1]$, and the eight expression probabilities are normalised. The complete 25-channel output enters numerical evaluation, while the avatar adaptor receives the 15 AU channels.

The main configuration follows the original facial frame count and audio duration, including the time-coordinate correction for the 25Hz text cache. Gaussian widths of five frames are used for speaker preprocessing, low-frequency output smoothing, and residual high-pass projection. Final smoothing uses 1.5 frames. These operations use the main facial grid. The 25Hz component protocols in Appendices~\ref{app:recording_generalisation},~\ref{app:vad_replacement}, and~\ref{app:diversity_audit} specify their own sampling grids and residual targets.

Each directed main-benchmark input uses one shared ITF trajectory and ten independently sampled facial predictions. The controlled Action-conditioned components use the generated Action semantic condition and their recorded affect configuration. Their predicted-guide training, target normalisation, AU-channel readout, and sampling counts are specified in the corresponding experiment protocols.

\subsection{Controlled Affect-Estimator Replacement}
\label{app:vad_replacement}

We evaluate replacement of the Action-text V/A estimator using frozen Qwen3-Embedding-0.6B representations with 1024 dimensions and Ridge regression. Human V/A/D labels from EmoBank provide supervision. Regularisation is selected on the official development set, followed by fitting on the combined 9062 training and development sentences. On the independent 1000-sentence test set, Pearson correlations for V/A/D are .777594, .535793, and .499881.

The facial component used for this experiment is trained from random initialisation for 60 epochs. Training Actions use observed training responses with word-level timing. The other state fields retain their automatic posterior records. Data preparation produces 2920 continuous windows and 23,014 aligned words. Of these, 2528 windows from 78 recordings provide fitting data, and 392 windows from 12 recordings provide checkpoint selection. Training uses learnt Action V/A, the original hand-crafted V/A-to-AU guide, and a 25-dimensional guide-error residual. Diffusion target statistics use fitting recordings.

At evaluation, we fix current Qwen3-8B-generated Actions, the 300-dimensional word-semantic interface, other state information, facial weights, paired diffusion seeds, and post-processing. Only the V/A readout from the successive Action prefixes changes. Four conditions use NRC, the frozen learnt estimator, retained State affect, and zero affect. The \texttt{None} action uses the zero-affect rule.

Generation uses 20 DDIM steps and CFG=3. Continuous AU outputs share the same range constraints and smoothing. Facial predictions are generated at 25Hz and resampled to each window's native frame clock. Low-frequency AU-MAE uses a Gaussian low-pass filter with a .5s time scale.

The primary confirmation contains 14 official validation recordings and 225 windows, excluding every original recording represented in the 102 official training groups. Each window fixes two generated Action texts. Each text and affect condition generates ten actual distinct facial trajectories, yielding 18,000 predictions.

A separate confirmation contains 12 recordings and 163 windows from a historical split within the official training pool. These recordings are excluded from fitting and checkpoint selection for the current component, yielding 13,040 predictions. The two confirmation groups are disjoint.

We average windows, generated texts, and facial samples within each recording before paired analysis. Two-sided Holm correction jointly covers AU-MAE, low-frequency AU-MAE, AU-CCC, variance gap, and jerk within each reported confirmation analysis.

\begin{table}[htbp]
\centering
\small
\caption{Affect-estimator replacement on 14 official validation recordings.}
\label{tab:vad_replacement}
\begin{tabular}{lrrr}
\toprule
Metric & NRC & Learnt & Holm $p$ \\
\midrule
AU-MAE & .257147 & .249118 & .034180 \\
Low-frequency AU-MAE & .217417 & .209793 & .049805 \\
AU-CCC & .000327 & .000502 & .781982 \\
Variance gap & .063407 & .063065 & .781982 \\
Jerk & .013391 & .012258 & .000610 \\
\bottomrule
\end{tabular}
\end{table}

Replacing NRC reduces overall AU error, low-frequency AU error, and jerk in the primary confirmation. CCC and variance gap improve numerically, with their adjusted comparisons remaining non-significant.

\begin{table}[t]
\centering
\scriptsize
\setlength{\tabcolsep}{2.5pt}
\caption{Additional confirmation and the prearranged recording-weighted pooled analysis. Each analysis corrects the same five trajectory comparisons.}
\label{tab:additional_confirmation}
\begin{tabular}{@{}llrrr@{}}
\toprule
Range & Metric & NRC & Learnt & Holm $p$ \\
\midrule
12 rec. & AU-MAE & .236767 & .226728 & .013672 \\
12 rec. & Low-freq. AU-MAE & .202229 & .192448 & .020508 \\
12 rec. & AU-CCC & .002839 & .000778 & .266113 \\
12 rec. & Variance gap & .056649 & .055711 & .084961 \\
12 rec. & Jerk & .013495 & .012347 & .002441 \\
\midrule
26 rec. & AU-MAE & .247741 & .238784 & .000281 \\
26 rec. & Low-freq. AU-MAE & .210407 & .201788 & .000652 \\
26 rec. & AU-CCC & .001487 & .000629 & .689320 \\
26 rec. & Variance gap & .060288 & .059671 & .066820 \\
26 rec. & Jerk & .013439 & .012299 & $1.49\times10^{-7}$ \\
\bottomrule
\end{tabular}
\end{table}

The separate 12-recording confirmation also supports lower overall and low-frequency AU errors and jerk. The prearranged pooled analysis gives equal weight to 26 disjoint recordings, comprising 388 windows and 31,040 facial trajectories. Its AU-MAE decreases from .247741 to .238784, and low-frequency AU-MAE decreases from .210407 to .201788. CCC decreases numerically in the 12-recording and pooled analyses.

\begin{table}[htbp]
\centering
\small
\caption{Complete AU-MAE controls for affect readout.}
\begin{tabular}{lrrrr}
\toprule
Range & NRC & Learnt & State & Zero \\
\midrule
14 recordings & .257147 & .249118 & .259636 & .247939 \\
12 recordings & .236767 & .226728 & .239388 & .225033 \\
26 recordings & .247741 & .238784 & .250291 & .237367 \\
\bottomrule
\end{tabular}
\end{table}

Zero affect has the lowest AU point error among these controls. Training uses learnt Action-affect conditions, so consistency with the training condition distribution forms part of this configuration. The confirmed result is the reduction in overall AU error, low-frequency AU error, and jerk when replacing NRC within this trained Action interface.

\begin{figure*}[htbp]
  \centering
  \includegraphics[width=\linewidth]{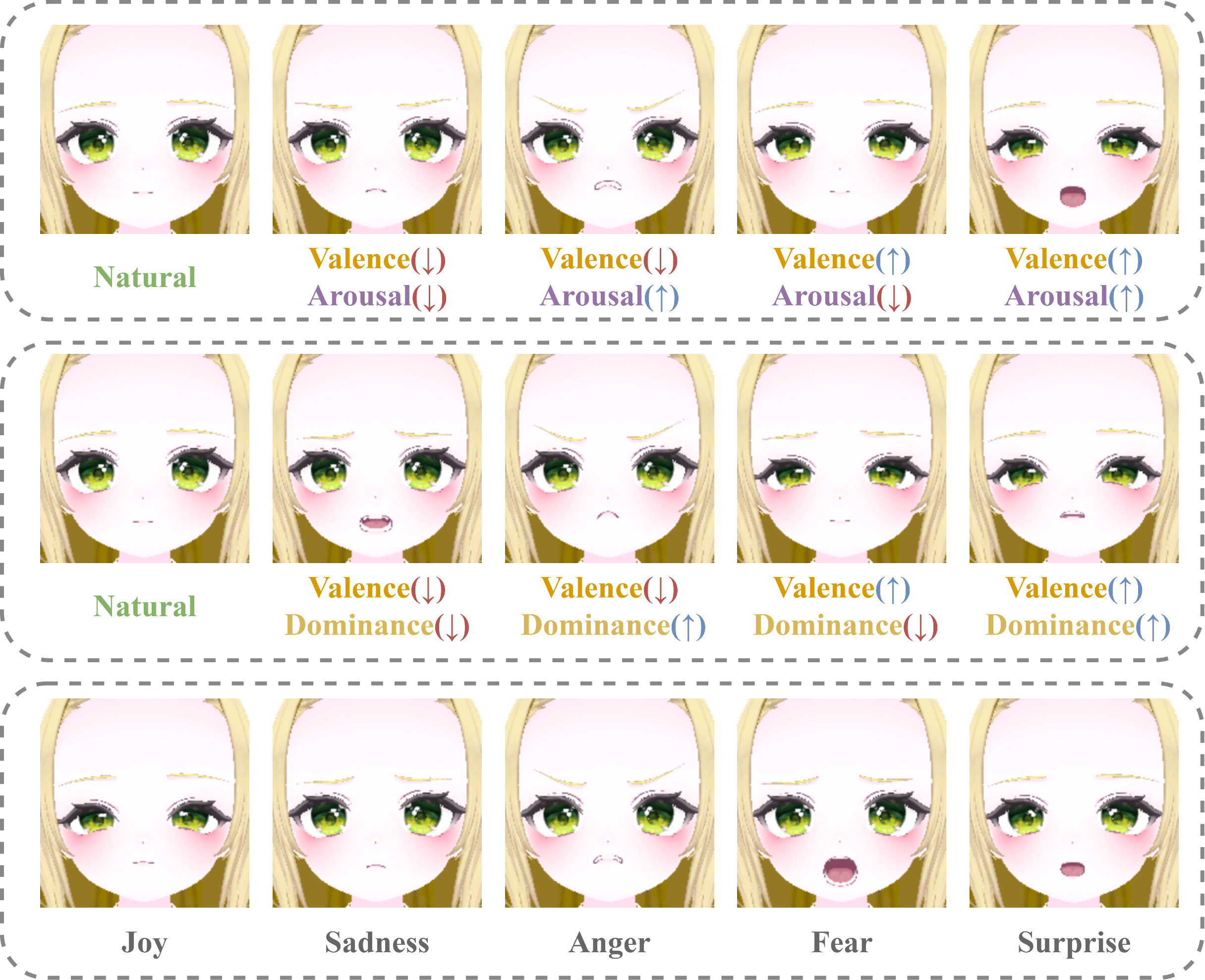}
  \caption{\textbf{Illustrative affective configurations of the avatar rig.} The panels show example configurations across valence--arousal and valence--dominance settings and discrete emotion categories.}
  \label{fig:facial_expression}
\end{figure*}

\section{Details of Experiments}\label{app:exp_setting}

\subsection{Dataset} 

Our primary evaluation uses the MARS benchmarking dataset \citep{song2025react} provided by the REACT 2025 challenge to evaluate the proposed intentional agent MAFRG. It contains 137 real human dyadic interaction clips involving 23 speakers and 137 listeners. These interactions ultimately constitute 270 multi-modal recording files. The original interactions last from 19 minutes and 50 seconds to 40 minutes and 49 seconds. During the data collection process, features such as audio and facial videos of each interacting speaker and listener pair were recorded independently. The entire dataset is divided into 2856 multimodal conversation pairs.

We retain natural speech gaps to evaluate reactions under discontinuous conversational timing. Speaker facial observations remain available in the natural-gap setting. Controlled probes specify the incoming information separately. Fixed-no-change probes clamp external perceptual content after initialisation, while the listener-task silence ablation supplies logical time and the simulated execution feedback defined by its protocol. The corresponding settings are detailed in Appendices~\ref{app:matched_itf_control} and~\ref{app:Controlled_Empirical_Validation}.

\subsection{Recording-Group-Disjoint Generalisation}
\label{app:recording_generalisation}

The component models use 78 fitting recording groups and 12 validation groups from the official training partition. They contain 2528 fitting windows and 392 validation windows. All model fitting, checkpoint selection, retrieval memories, and empirical target banks exclude the confirmation recordings. Confirmation uses every eligible speaker-input window from 14 official test recording groups absent from all official training and validation recordings. The input-only speech-offset rule yields 242 windows of 256 native facial frames and excludes 52 inputs. Source audio durations determine the temporal alignment.

This study uses an Action-conditioned component configuration. Paired training inputs and responses supply automatic posterior state hypotheses. The low-frequency mapping takes 25 speaker attributes, five retained V/A terms, and ten acoustic features. Speaker and thought semantics each contribute 32 projected features, giving 104 inputs. The facial DiT receives 300 Action-semantic features, two State V/A features scaled by three and two, and 15 predicted guide AUs. Its causal backbone uses width 256, eight heads, and eight layers.

Let $Y$ denote the aligned 25-channel training trajectory and $H$ the predicted guide. Full predicts the correction $U^{\rm F}=Y-H$, and Learnt-Only predicts $U^{\rm L}=Y$ with a zero physical guide. Each target uses fitting-only channel means and standard deviations. Both models use random initialisation, 60 epochs, batch size 64, and training seeds 20261111, 3408, and 3409. The training objective combines noise MSE with twice the CCC loss.

The 100-step linear noise schedule uses endpoints $10^{-4}$ and $.02$, with terminal cumulative signal coefficient $\bar\alpha_T=.36356$. To initialise sampling, we draw a fitting target from an empirical bank, standardise it, and apply forward noise
\[
x_T=\sqrt{\bar\alpha_T}\,\widetilde U_j
    +\sqrt{1-\bar\alpha_T}\,\epsilon,
\qquad \epsilon\sim\mathcal N(0,I).
\]

The two banks index the same 2516 complete eight-second fitting windows. Full uses standardised corrections, and Learnt-Only uses standardised full targets. The sampled sequence is interpolated to the current temporal grid. Decoding uses 20 DDIM steps and CFG=3 on the 25Hz grid.

Full adds the generated 15-channel AU correction to $H$ and applies 1.5-frame Gaussian smoothing. Predictions return to the native frame grid, where AU occurrences use threshold .5. Each input uses two speaker-derived state flows, assigned equally across ten independently seeded facial samples. The six neural models generate 14,520 trajectories, and the mapping control contributes 2420 realisations.

Mapping-Only retains $H$. The Mapping + clean prior control adds the sampled clean fitting correction directly to $H$. Full starts from the forward-noised target and runs conditional denoising. The clean-prior control retains the same guide, empirical source selection, inverse normalisation, and output processing. It performs zero neural denoising steps and contributes 2420 template draws.

VaG and VeG follow the existing implementation on the concatenated speaker-silent AU frames. They measure absolute differences in mean temporal variance and mean absolute first differences. DG uses the complete window. It measures the absolute difference between the speaker--prediction and speaker--reference DTW costs of z-scored AU-motion envelopes, downsampled by five. The gaps quantify agreement in AU-motion statistics and DTW costs. Statistics first average all windows and repeated model and sampling seeds within each recording, then apply two-sided recording-level paired tests.

\begin{table*}[htbp]
\centering
\caption{Recording-group-disjoint confirmation on 14 REACT 2025 test recordings and 242 speaker inputs. Lower values are better.}
\label{tab:recording_generalisation}
\begin{tabular}{lrrr}
\toprule
Condition & VaG & VeG & DG \\
\midrule
Full & \textbf{.039641} & \textbf{.019242} & \textbf{.103148} \\
Learnt-Only & .042819 & .021713 & .113554 \\
Mapping-Only & .055495 & .029073 & .149437 \\
Mapping + clean prior & .041258 & .021121 & .107036 \\
\bottomrule
\end{tabular}
\end{table*}

The three gaps against Learnt-Only and Mapping-Only form six primary comparisons. Their joint Holm-adjusted $p$ values are at most .003052. Full has the lowest VaG, VeG, and DG among the four conditions. The three supplementary comparisons against Mapping + clean prior form a separate Holm family, with adjusted $p=.010498$, $.001099$, and $.013428$ for VaG, VeG, and DG. These results support the combined guide and conditional generation branch on the held-out recordings.

Companion AU-MAE values are .245883 for Full, .263770 for Learnt-Only, .195629 for Mapping-Only, and .254003 for Mapping + clean prior.

The main-checkpoint evaluation excludes all recordings represented in its training and validation materials. Its subset contains 588 directed inputs from 14 recordings, with ten predictions per input. Clip-weighted FRCorr, FRDist, FRDiv, FRVar, and FRSyn are .78461, 117.90384, .01775, .02273, and 47.36395. Recording-weighted values are .78420, 118.25106, .01777, .02273, and 47.37032.

\subsection{Diversity Source and Plausibility Analysis}
\label{app:diversity_audit}

We use a separately trained, channel-normalised Full facial component. It is trained from random initialisation for 60 epochs with a 100-step training noise schedule. Generation uses 20 DDIM steps and CFG=3. Neural conditions use Action semantics and the agent's retained affect. The low-frequency interface and retrieval memories are fitted from automatic posterior state records constructed from training input and response pairs. The low-frequency interface uses 104 features, comprising the 72 speaker-and-affect features and 32 additional projected features from the current Thought. The DiT's 300-dimensional semantic condition uses the generated Action text.

Confirmation uses 14 official validation recordings disjoint from all 102 official training recordings. They are also disjoint from the 14 recordings reserved for the separate diversity analysis. The protocol is first examined on another 12 development recordings. For each confirmation recording, we take the first eligible speaker window in fixed source order, extract eight seconds, and resample to 25Hz.

Each recording produces ten independent state trajectories with four updates each, yielding 140 state trajectories and 560 actual updates. State-Only varies the state trajectory while fixing the diffusion initial value. Noise-Only fixes the first state trajectory and varies the diffusion initial value. The Both condition varies both sources. Each condition generates ten distinct facial trajectories per recording, yielding 420 trajectories in 42 prediction stacks.

\begin{table}[htbp]
\centering
\caption{Sources of variation in actual ten-candidate generation. FRDiv and FRVar use the 25-dimensional outputs.}
\label{tab:diversity_sources}
\begin{tabular}{lrr}
\toprule
Condition & Mean pairwise difference & FRVar \\
\midrule
Noise-Only & .023660 & .015442 \\
State-Only & .010362 & .016704 \\
Both & .032964 & .016412 \\
\bottomrule
\end{tabular}
\end{table}

Both produces the largest mean pairwise difference. In the 15 AU channels, State-Only assigns 78.88\% of candidate-difference energy below .5Hz and 7.50\% above 2Hz. Noise-Only assigns 36.76\% and 22.78\%, respectively.

Motion analysis uses continuous AU commands in $[0,1]$. We match the mean pairwise MSE of the 15 AU channels using a common centre trajectory and common physical bounds. All 126 combinations of three amplitudes, three sources, and 14 recordings reach the target. Speed, jerk, and jumps use discrete frame differences at 25Hz. Boundary proximity is the proportion of AU values below .05 or above .95.

\begin{table}[htbp]
\centering
\small
\caption{Trajectory characteristics at matched AU pairwise MSE .01333333. Spectral proportions are measured before amplitude matching.}
\label{tab:diversity_structure}
\begin{tabular}{lrr}
\toprule
Measure & State-Only & Noise-Only \\
\midrule
Out-of-range proportion & .000000 & .000000 \\
Boundary proximity & .442331 & .444850 \\
Speed per frame & .019806 & .024111 \\
Jerk per frame & .017645 & .019862 \\
99th-percentile jump & .114376 & .140361 \\
Maximum jump & .254711 & .281752 \\
Energy below .5Hz & 78.88\% & 36.76\% \\
Energy above 2Hz & 7.50\% & 22.78\% \\
\bottomrule
\end{tabular}
\end{table}

We jointly apply two-sided Holm correction to nine predefined comparisons, covering jerk, 99th-percentile jump, and maximum jump at three target amplitudes. Original recordings are the independent units.

\begin{table*}[htbp]
\centering
\small
\caption{Matched-amplitude motion comparisons. Each measure is lower for State-Only. All nine tests share one Holm correction family.}
\begin{tabular}{llrrr}
\toprule
Target AU pairwise MSE & Measure & State-Only & Noise-Only & Holm $p$ \\
\midrule
.00333333 & Jerk & .013955 & .014815 & .001099 \\
.00333333 & 99th-percentile jump & .091557 & .099769 & .001099 \\
.00333333 & Maximum jump & .173827 & .200936 & .005127 \\
\midrule
.00666667 & Jerk & .015303 & .016693 & .003052 \\
.00666667 & 99th-percentile jump & .099367 & .114809 & .001099 \\
.00666667 & Maximum jump & .202801 & .233339 & .008057 \\
\midrule
.01333333 & Jerk & .017645 & .019862 & .003418 \\
.01333333 & 99th-percentile jump & .114376 & .140361 & .001099 \\
.01333333 & Maximum jump & .254711 & .281752 & .049438 \\
\bottomrule
\end{tabular}
\end{table*}

All nine adjusted comparisons favour State-Only. Both sources have zero out-of-range proportion after common bounded processing. Boundary proximity differs numerically, with its comparison remaining non-significant. These results support slower expression trends from state variation and smoother continuous AU motion at matched candidate-difference amplitudes.

As a descriptive robustness check, we remove the two recordings with the highest Both-condition FRDiv, corresponding to 14.29\% of recordings. The same remaining 12 recordings are used for all sources. Mean pairwise differences are .009359 for State-Only, .022961 for Noise-Only, and .031559 for Both. Both remains the most diverse condition.

This experiment uses the stated continuous-command component model. The formal REACT main table uses its corresponding saved p55 evaluation configuration.

\subsection{Cross-Benchmark Adaptation on Seamless Interaction}
\label{app:seamless_adaptation}

We evaluate modular adaptation on \textit{Seamless Interaction}~\citep{seamless_interaction}. The state-update model, prompts, low-frequency mapping, and acoustic missing-modality rules remain fixed. Only the facial DiT is adapted.

Four training windows provide adaptation data, and one validation window selects checkpoints and the shared decoder. Training conditions use actual speaker attributes, transcripts, and independently inferred ITF. Residual targets subtract the corresponding inference-produced low-frequency base from the target reaction.

Source and adapted models share a mean-preserving decoder, high-frequency limits, smoothing, and physical bounds. Confirmation uses 12 project-held-out sessions, three adaptation seeds, and ten predictions per input. These sessions come from the official training pool and were excluded from this project's fitting and selection.

\begin{table}[htbp]
\centering
\caption{Seamless adaptation under a common mean-preserving decoder.}
\label{tab:seamless_adaptation}
\begin{tabular}{lrrr}
\toprule
Condition & AU-MAE & Velocity gap & AU-CCC \\
\midrule
Source Full & .230341 & .009483 & .003897 \\
Adapted Full & .100567 & .004601 & .007911 \\
Training mean & .085028 & .004326 & .000000 \\
Training median & .065903 & .004326 & .000000 \\
\bottomrule
\end{tabular}
\end{table}

Adapting only the DiT reduces AU-MAE from .230341 to .100567 and velocity gap from .009483 to .004601. Both primary session-level comparisons have Holm-adjusted $p=.000977$. The AU-CCC comparison gives $p=.30127$. These results support target-domain adaptation of parameter distributions and movement amplitude relative to the source model.

\subsection{Implementation Details}\label{detail_ID}

The agent receives time-stamped textual and numerical descriptions of the input speaker. Whisper supplies speech transcripts and word times. Word intervals are mapped to facial frames using audio duration and original frame count. Each state-generation observation bin contains speaker words, mean facial V/A values, and mean occurrences of the 15 AUs. The low-frequency facial module also receives frame-level speaker attributes and acoustic features.

The retained main configuration uses four-second observation bins. Offline memory retrieval queries the complete input transcript and supplies an analogous official-training interaction with observed listener responses. Up to three recent states are included at each update. The Qwen3-8B Q4\_K\_M uses an 8192-token context, with its model parameters fixed during inference. The current prompt specifies action-progression instructions under unchanged input, including progression after a None action and physical movement after three consecutive no-change waits. NRC-derived appraisals participate in the programmed affect update.

The newer listener-task protocols use the observation bins and update times specified by each experiment. The controlled silence ablation updates at 1, 2, and 3s after silence onset. Contextual variants use the frozen affect estimator and their recorded appraisal-text scope. Training memories contain automatic posterior state hypotheses inferred from observed training input--response pairs. These hypotheses provide retrieval examples and facial-training conditions. Retrieval and affective initialisation use the designated fitting recordings. The listener-task prompt permits continued listening and treats \texttt{None} as continuation without a new command. Execution feedback follows each experiment's protocol.

The affect-estimator experiment aligns observed training responses at word level. The diversity component uses Action text for its DiT semantic condition, with token vectors distributed within state intervals. Its low-frequency interface additionally uses projected Thought features. Generated Action tokens are placed within their state intervals at inference. Formal main-benchmark inference uses the 72-feature low-frequency model, the retained 25-channel residual readout, 20 DDIM steps, and CFG=3.

\paragraph{Runtime and Interactive Feasibility.}
\label{app:runtime_comparison}

Measurements use an RTX 5090 Laptop GPU and Qwen3-8B Q4\_K\_M. The complete path includes memory retrieval, affect updating, state generation, parsing, condition construction, facial generation, and display submission. Eight contexts are evaluated three times after warm-up. Each display update generates one 200-frame trajectory. Timing starts at perceived-text availability and ends after two requestAnimationFrame callbacks in headless Chromium using SwiftShader. ASR processing precedes this boundary.

\begin{table*}[htbp]
\centering
\caption{Post-perception update and headless-browser submission latency.}
\begin{tabular}{lrrr}
\toprule
Configuration & Runs & Mean (s) & P95 (s) \\
\midrule
Full, 20 DDIM steps & 24 & 1.9645 & 2.3510 \\
Full, 3 DDIM steps & 24 & 1.7936 & 2.1880 \\
PerFRDiff & 30 & 2.7610 & 2.8390 \\
\bottomrule
\end{tabular}
\end{table*}

For the formal 20-step configuration, mean state-update time is 1.6987s and numerical facial processing takes .1921s. Numerical readiness is reached after 1.8917s. The 3-step development configuration has corresponding means of 1.6861s, .0330s, and 1.7201s.

The PerFRDiff reference retains its earlier protocol, beginning from precomputed-feature availability. It uses the authors' generic weights, encoders, 50 prior steps, and 50 decoder steps. Computational inputs include real speaker facial attributes and fixed-size synthetic audio, 3DMM, and individual-reference tensors.

A separate 60-run, 200-frame, 20-step DiT microbenchmark takes .4310s on average, with P95 .4534s.

\paragraph{Event scheduling.}
Across eight rapid events, mean cancellation-request delay is 34.71ms and P95 is 54.47ms. The final event reaches numerical readiness after 2.4167s and browser submission after 2.4632s. With a 15s idle heartbeat, a new event starts state inference after .1333s, reaches numerical readiness after 2.6454s, and reaches browser submission after 2.7026s.

\paragraph{Sampling-step development curve.}
An eight-recording development experiment uses ten candidates per input. Mean generation times for ten candidates are .5316s, .6969s, 1.0265s, and 1.4929s for 3, 6, 10, and 20 steps. Their FRCorr values are .97970, .97892, .97900, and .97927. Formal main evaluation uses 20 steps.

\paragraph{Additional Runtime Measurements.}
The latest model configuration averaged 4.29s for state updating and 4.37s for the complete update. Its DiT-only microbenchmark generated 375 frames in .079s with six diffusion steps. The 30-minute evaluation had a peak RSS of 2.09GB and memory growth of 10.1MB. All measurements were obtained using the specified current configuration and consistent measurement boundaries.

\subsection{Metrics}

\paragraph{Statistical analysis.}
We use the independent units specified by each experiment and aggregate repeated seeds within those units. State-role analysis applies Holm correction to three predefined comparisons. The silence-channel experiment jointly corrects seven primary comparisons, covering four command-rate comparisons and three self-reported expectation-error comparisons. Free-thought continuity applies two-sided Holm correction to two primary endpoints, with Only-Thought comparisons reported as supplemental controls. Matched intention retention and terminal facial readout each correct two control comparisons per endpoint. Affect-estimator replacement corrects five trajectory metrics within each confirmation cohort and the prearranged pooled analysis. Recording-group-disjoint confirmation jointly corrects six comparisons across three gaps and two original component controls. Its three comparisons with the clean empirical-prior control form a separate Holm family. Diversity analysis jointly corrects nine two-sided comparisons across three amplitudes and three motion measures. The auxiliary V/A interface retains the complete 45-comparison Holm correction.

\subsubsection{Facial Reaction Metrics}

We adopted the multidimensional evaluation framework defined by REACT 2025 to evaluate the listener emotional reactions generated by the model. This framework primarily evaluates the generated facial emotional reactions from the following three core dimensions:

\textbf{Appropriateness.} FRCorr uses the concordance correlation coefficient (CCC) to compare generated and reference facial reactions. For each generated reaction, it selects the highest CCC over the available reference reactions. FRDist uses dynamic time warping (DTW) and selects the lowest weighted distance over these references. Both metrics sum the selected values across generated reactions and average them across inputs.

\textbf{Diversity.} This measures the richness of the reactions generated by the model when facing different speaker inputs or the same input. It is quantified through two metrics, FRVar and FRDiv.

\textbf{Synchrony.} This evaluates the temporal alignment relationship between the generated facial reactions and the speaker behaviours. It is measured using time-lagged cross-correlation (TLCC) and is denoted as FRSyn in the standard.

\subsubsection{Action Evaluation Metrics}

To quantify the action performance of the agent in complex interactive scenarios, we designed an automated evaluation framework based on the LLM-as-a-Judge approach, utilising real human interaction data included in REACT 2025. MARS supplies natural conversational gaps for contextual reaction evaluation. Controlled zero-input experiments initialise the agent from a recorded context and then stop new perceptual content according to Appendix~\ref{app:matched_itf_control}. All evaluation dimensions are quantitatively scored using a 1 to 5 Likert scale. A score of 1 represents extremely poor or completely non-compliant, while a score of 5 represents excellent or fully compliant. Specifically, we comprehensively examine the following five core dimensions:

\textbf{Affect-Action Coherence.} This evaluates whether the emotional state parameters generated at the foundational level of the agent are logically mapped in its explicit text responses and pragmatic features.

\textbf{Pragmatic Utility.} This evaluates whether the response of the agent achieves a positive and meaningful communication function at the current conversation node.

\textbf{Temporal Intentionality.} This evaluates whether the temporal dynamic characteristics of the agent during interaction align with human physiological and cognitive norms when performing corresponding complex cognitive tasks.

\textbf{Role Compliance.} This evaluates whether the agent strictly adheres to its assigned system identity boundaries throughout long texts and multi-turn interactions.

\textbf{Affective Appropriateness.} This evaluates whether the overall emotional tone revealed by the agent aligns with the current scenario when executing specific interaction strategies.

\subsection{Additional Human-Evaluation Protocol}
\label{app:additional_human_protocol}

\paragraph{Study design and presentation.}
The first additional study contains 12 contexts and four conditions, yielding 48 unique stimuli. Six completed participants provide 288 records. The original Full is compared with B\_Mime, and the separately trained component Full is compared with Learnt-Only. The latter pair shares Agent conditions, training budget, training and diffusion seeds, and post-processing.

The second study combines 24 ITF-control contexts and 12 external-comparison contexts. Four completed participants each evaluate 24 clips, yielding 96 records over 72 unique stimuli. ITF contributes 64 records and 32 participant--context pairs. The external comparison contributes 32 records and 16 pairs. Twelve contexts have two raters and 24 have one rater. All 36 contexts are covered.

The interface supplies the dialogue context and method-anonymous avatar clips. Paired conditions use the same avatar, camera, and AU amplitude with additional idle animation disabled. The ITF material follows Appendix~\ref{app:matched_itf_control}. The external condition uses CaReDiff's published generic offline weights with matching real speaker facial attributes and the authors' native None path for audio and 3DMM. This condition is labelled CaReDiff face-attribute input.

\paragraph{Inclusion and analysis.}
All completed participants are included. Playback duration is retained as process information. Unrateable responses are treated as missing for paired scoring. Differences are computed within each participant and context, then averaged over participants within each context. Two-sided Wilcoxon tests use contexts as units. Each study's two comparisons and two outcomes form a four-test Holm family. Confidence intervals use 5,000 context-bootstrap samples.

\begin{table*}[htbp]
\centering
\small
\caption{\textbf{Context-bootstrap 95\% intervals for the additional human comparisons.}}
\begin{tabular}{lll}
\toprule
Comparison & Context appropriateness & Naturalness \\
\midrule
Full (orig.)--B\_Mime & [1.347, 2.250] & [1.347, 2.333] \\
Full (comp.)--Learnt-Only & [.694, 1.236] & [.319, .847] \\
Full (ITF)--Factual Summary & [.542, 1.188] & [.667, 1.292] \\
Full (orig.)--CaReDiff-face & [.792, 2.042] & [.625, 1.500] \\
\bottomrule
\end{tabular}
\end{table*}

\paragraph{Participant and context sensitivity.}
All six first-study participants favour original Full over B\_Mime on both outcomes. Contextual appropriateness is positive in all 12 contexts, and naturalness is positive in 11 with one tie. Independently resampling participants and contexts gives intervals of [.819, 2.778] for contextual appropriateness and [.764, 3.014] for naturalness. For component Full versus Learnt-Only, the corresponding exploratory intervals are [.291, 1.736] and [-.069, 1.278].

For the four-participant study, 20,000 independent participant-and-context resampling attempts give ITF-control intervals of [.267, 1.533] and [.188, 1.667], and external-comparison intervals of [.250, 2.286] and [.125, 1.917], respectively. Sparse resamples average available paired cells. The external outcomes have 19,986 and 19,988 valid resamples. These checks are reported separately from the primary context-level tests.

\paragraph{Exploratory learnt-branch comparison.}
Learnt-Only exceeds B\_Mime by 1.458 points in contextual appropriateness and 1.319 in naturalness. Including this comparison with the two first-study comparisons yields six jointly Holm-corrected tests, with $p=.002930$ for both added outcomes. Participant-and-context bootstrap intervals are [.653, 2.333] and [.486, 2.264].

\subsection{Metric and Motion Diagnostics}
\label{app:metric_diagnostics}

\paragraph{Reference sensitivity.}
Fourteen raw human listener trajectories produce 91 pairwise comparisons over common temporal prefixes. Mean CCC is .013893, median CCC is .010488, and 93.41\% of pairs have CCC below .1. These comparisons share source trajectories and are descriptive.

For fixed predictions on 14 clips with ten candidates per clip, single-reference FRCorr is .648298, the original full-reference-pool value is 1.079144, and the official ten-reference value is 1.007582. On the 559-clip diagnostic subset using the first three actual predictions per input, the corresponding values are .189853, .331383, and .266184. FRCorr aggregates the best-reference CCC across generated candidates, making candidate count and reference construction part of its reported protocol.

\paragraph{Candidate variation and temporal motion.}
The 14-clip ten-candidate diagnostic has mean pairwise difference .021624, temporal variance .030040, and mean absolute frame-to-frame AU change .018655. The 559-clip three-candidate diagnostic has corresponding values of .020269, .027621, and .017628.

\paragraph{Constant channels and synchrony.}
On the diagnostic data, original TLCC gives 44.7943 frames using the first prediction and 44.7394 frames using the candidate-mean prediction. The finite-channel diagnostic gives 8.4562 and 8.2952 frames, respectively. The formal main FRSyn result uses the original implementation and first prediction.

\paragraph{Fixed-state residual-gain intervention.}
We hold state conditions, checkpoint, diffusion noise, and post-processing fixed over 12 contexts.

\begin{table*}[htbp]
\centering
\caption{Effect of residual gain on AU motion and raw out-of-range values.}
\begin{tabular}{rrrr}
\toprule
Gain & AU speed & Temporal variance & Raw out-of-range proportion \\
\midrule
0 & .010619 & .017861 & .397222 \\
1 & .016345 & .019189 & .387250 \\
10 & .086527 & .075116 & .500778 \\
100 & .172512 & .205442 & .822417 \\
\bottomrule
\end{tabular}
\end{table*}

The current default residual gain is one. Increasing gain raises motion magnitude and raw out-of-range values. Residual gain and AU-probit scale .55 are separate readout parameters. These diagnostics establish the effects of reference construction, constant-channel handling, and readout gain on metric interpretation.

\subsection{Action-Level Diagnostics Across Foundation Models}
\label{app:action_diagnostics}

We evaluate the first non-empty action generated by current Qwen3-8B for each of 14 session0 videos. GPT-5.2 applies the original judging prompt on five 1--5 criteria. Real listener speech is supplied to the judge during scoring. Other backbone rows retain their original generation and judging versions.

\begin{table*}[htbp]
\centering
\caption{Action-level automatic scores.}
\label{tab:semantic_evaluation}
\begin{tabular}{lccccc}
\toprule
Model & \makecell{Affect-Action\\Coherence} & \makecell{Pragmatic\\Utility} & \makecell{Temporal\\Intentionality} & \makecell{Role\\Compliance} & \makecell{Affective\\Appropriateness} \\
\midrule
GT & 3.00 & 1.93 & 2.43 & 3.07 & 2.64 \\
Qwen3-8B & 3.79 & 3.21 & 3.86 & 4.64 & 4.29 \\
Llama3.1-8B & 3.21 & 3.07 & 3.36 & 3.43 & 3.79 \\
GPT-4o & 2.71 & 2.43 & 3.50 & 3.36 & 3.43 \\
DeepSeek-V3 & 3.29 & 2.36 & 3.79 & 3.86 & 3.79 \\
\bottomrule
\end{tabular}
\end{table*}

The current Qwen3-8B means are 3.79 for affect-action coherence, 3.21 for pragmatic utility, 3.86 for temporal intentionality, 4.64 for role compliance, and 4.29 for affective appropriateness. These scores provide an auxiliary diagnostic of generated actions.

\section{Details of the Reaction Quality Scorer}
\label{app:rqs_details}

To ensure complete reproducibility of our proposed Reaction Quality Scorer (RQS), we provide detailed descriptions of the model architecture, data processing, and the dynamic negative sampling strategy utilised during training.

\subsection{Model Architecture}
The RQS is built upon a Cross-attention Transformer architecture designed to model the temporal correlation between the speaker's stimuli and the listener's reactions.

    \paragraph{Projection and Positional Encoding.} The raw 15-dimensional AU sequences of both the listener and speaker are first projected into a hidden dimension of $D=128$ via linear layers. A learnable positional encoding (\texttt{nn.Embedding}) is added to both sequences.
    \paragraph{CLS Token and Self-Attention.} A learnable \texttt{[CLS]} token is prepended to the listener's sequence. The listener's sequence then passes through a 2-layer Transformer Encoder (num\_heads=4, dim\_feedforward=256) to model the internal temporal dynamics of the listener's facial expressions.
    \paragraph{Cross-Attention.} The encoded listener sequence acts as the Query (\textit{tgt}), while the speaker's sequence acts as the Key and Value (\textit{memory}). This interaction is modeled using a 2-layer Transformer Decoder, explicitly conditioning the listener's reaction on the speaker's context.
    \paragraph{Scoring Head.} The output representation corresponding to the \texttt{[CLS]} token is extracted and passed through a Multi-Layer Perceptron (MLP) head: \texttt{LayerNorm $\rightarrow$ Linear(128, 64) $\rightarrow$ GELU $\rightarrow$ Dropout(0.1) $\rightarrow$ Linear(64, 1) $\rightarrow$ Sigmoid}. This ensures the final output is a scalar score $s \in [0, 1]$.

\subsection{Dynamic Online Negative Sampling and Label Construction}
To prevent the model from trivially distinguishing human faces from noise, and to force it to learn precise semantic and temporal alignments, we employ a dynamic online sampling strategy at the dataset level. For each epoch, we construct a batch with a specific ratio of Ground Truth to synthesised negative samples (Ratio = 1.0 : 1.0 : 1.0 : 0.5).

Given a ground truth pair $(L, S)$, the augmented samples and their soft labels are constructed on-the-fly as follows:

\paragraph{Ground Truth (Label $1.0$).} The original aligned pair $(L, S)$.
\paragraph{Time-Shifted GT (Label $0.85 \sim 0.90$).} The listener's sequence $L$ is temporally shifted by a random continuous duration $\Delta t \in [0.5, 1.0]$ seconds. To reflect human tolerance to slight delays, the quality score is penalised linearly: $s_{shift} = 0.95 - 0.1 \times \Delta t$.
\paragraph{Mismatched Context (Label $0.20 \sim 0.40$).} The listener's sequence $L$ is paired with a speaker sequence $S_{other}$ randomly sampled from a different interaction session. The score is uniformly sampled from $\mathcal{U}(0.2, 0.4)$, acknowledging the physical realism of the face but penalising the semantic mismatch.
\paragraph{Gaussian Noise (Label $0.0$).} $L$ is replaced with random Gaussian noise multiplied by a factor of $2.0$. This acts as a regulariser to anchor the lower bound of the score. All RQS training samples are constructed from real speaker-listener pairs in REACT 2025 and the perturbations above. Outputs from our facial reaction generator are not used during RQS training. RQS therefore does not directly learn output patterns specific to our generator, although it remains trained within the REACT 2025 distribution.

\subsection{Training Setup}
The model is trained using the AdamW optimiser with a learning rate of $3 \times 10^{-4}$ and a weight decay of $10^{-4}$. We utilise a \texttt{CosineAnnealingLR} scheduler ($T_{max}=500, \eta_{min}=10^{-6}$). To mitigate gradient explosion, gradient clipping is applied with a max norm of $1.0$. The model is optimised using a Smooth L1 Loss ($\beta=0.1$) to provide robustness against the noise introduced by the continuous soft labels. The training batch size is set to $64$.

\subsection{Session-Level RQS Results}
\label{app:rqs_session_results}

Table~\ref{tab:rqs_session_results} reports the complete session-level RQS evaluation on the REACT 2025 test set.

\begin{table*}[htbp]
\centering
\caption{\textbf{Complete RQS evaluation results across conversational sessions.}}
\label{tab:rqs_session_results}

\begin{tabular}{l|cccc|cc}
\toprule
\textbf{Category} & \textbf{Mean} & \textbf{Median} & \textbf{Std} & \textbf{Samples} & \textbf{GT Score} & \textbf{Relative Ratio (\%)} \\
\midrule
\textit{Overall} & $0.705$ & $0.854$ & $0.242$ & $559$ & $0.854$ & $82.6$ \\
\midrule
Session 0 & $0.507$ & $0.437$ & $0.222$ & $14$ & $0.756$ & $67.2$ \\
Session 1 & $0.627$ & $0.637$ & $0.191$ & $14$ & $0.872$ & $71.9$ \\
Session 2 & $0.806$ & $0.889$ & $0.158$ & $112$ & $0.886$ & $91.0$ \\
Session 3 & $0.570$ & $0.591$ & $0.197$ & $14$ & $0.857$ & $66.6$ \\
Session 4 & $0.660$ & $0.725$ & $0.215$ & $14$ & $0.906$ & $72.8$ \\
Session 5 & $0.854$ & $0.862$ & $0.050$ & $14$ & $0.874$ & $97.7$ \\
Session 6 & $0.911$ & $0.911$ & $0.000$ & $2$ & $0.903$ & $100.8$ \\
Session 7 & $0.593$ & $0.592$ & $0.236$ & $14$ & $0.902$ & $65.7$ \\
Session 9 & $0.868$ & $0.888$ & $0.053$ & $14$ & $0.867$ & $100.2$ \\
Session 10 & $0.893$ & $0.909$ & $0.070$ & $113$ & $0.899$ & $99.3$ \\
Session 11 & $0.856$ & $0.902$ & $0.110$ & $13$ & $0.852$ & $100.4$ \\
Session 12 & $0.851$ & $0.900$ & $0.119$ & $13$ & $0.894$ & $95.2$ \\
Session 13 & $0.532$ & $0.498$ & $0.187$ & $14$ & $0.826$ & $64.4$ \\
Session 14 & $0.472$ & $0.460$ & $0.201$ & $14$ & $0.810$ & $58.2$ \\
Session 15 & $0.416$ & $0.366$ & $0.191$ & $14$ & $0.836$ & $49.7$ \\
Session 16 & $0.464$ & $0.359$ & $0.229$ & $14$ & $0.844$ & $55.0$ \\
Session 17 & $0.337$ & $0.287$ & $0.166$ & $14$ & $0.833$ & $40.5$ \\
Session 18 & $0.546$ & $0.482$ & $0.249$ & $111$ & $0.768$ & $71.1$ \\
Session 19 & $0.745$ & $0.836$ & $0.161$ & $14$ & $0.895$ & $83.3$ \\
Session 21 & $0.881$ & $0.888$ & $0.025$ & $13$ & $0.903$ & $97.6$ \\
\bottomrule
\multicolumn{7}{l}{\small Relative Ratio = Model Score / GT Score $\times$ 100\%.}
\end{tabular}

\end{table*}

The session-level scores vary substantially across conversational contexts. For example, the relative ratio reaches $100.8\%$ in Session 6, whereas Session 17 obtains $40.5\%$. We therefore use the aggregate RQS result only as supplementary automatic evidence and validate its agreement with human judgements separately in Section~\ref{sec:rqs_human}.

\subsection{Recording-Group-Disjoint Validation of RQS}
\label{app:rqs_group_disjoint}

To test whether the human correlation could result from direct overlap with the evaluated recording groups, we retrained RQS after excluding all 14 recording groups represented by the 96 human-rated stimuli from both training and validation. The remaining training and validation sets contained 82 and 20 recording groups, respectively. Model selection did not access the excluded stimuli or their human scores.

On the 96 human-rated stimuli, the retrained RQS still correlated with human scores at Pearson $r=.744$ and Spearman $\rho=.788$. After removing condition-level mean differences, the corresponding correlations were $r=.412$ and $\rho=.383$. After removing the mean difficulty of each context, the corresponding correlations were $r=.770$ and $\rho=.676$. Within each context, RQS agreed with human pairwise rankings for $81.94\%$ of reaction pairs.

These results show that the observed human agreement persists when the evaluated recording groups are excluded from RQS training and validation, and therefore is not explained by direct memorisation of those recording groups. Because the retrained scorer still uses REACT 2025 data, this experiment does not remove all dataset-level priors or establish cross-dataset agreement with human judgement. We therefore continue to use RQS as a supporting metric alongside human evaluation and trajectory-based metrics.

\subsection{Human Evaluation and Validation of RQS}
\label{app:rqs_human}

\paragraph{Internal comparison conditions.}
Event-Triggered is constructed at the facial-output level. It uses the Full trajectory when speaker input is present and holds the last AU frame during speech gaps, followed by the common smoothing and range constraints. Heuristic-Only retains the rule-based V/A-to-AU guide and removes the DiT residual. Full adds the checkpoint-scaled residual to that guide before post-processing. These conditions test output continuity and the learnt facial contribution. The matched-update thought-carry intervention separately tests retained state content in Appendix~\ref{app:matched_itf_control}.

We recruited eight raters from online communities to assess 96 facial reactions from 24 settings and four conditions: GT, Full, Event-Triggered, and Heuristic-Only. GT is the real human reaction. Event-Triggered uses the Full trajectory when speaker input is present and holds the last AU frame during speech gaps. The common smoothing and range constraints are then applied. Heuristic-Only uses the hand-written V/A-to-AU mapping rules of that study's facial implementation. The raters knew neither the condition names nor the RQS scores. Statistical tests used the 24 settings as paired units with Holm-corrected Wilcoxon tests.

The raters used 7-point Likert scales to assess naturalness, context fit, emotion fit, temporal continuity, and credibility. Inter-rater consistency was high: ICC(2,8) ranged from $.965$ to $.978$ across the five measures, while the total-score ICC was $.986$.

Full had the highest mean human score at $5.527$ and did not differ significantly from GT at $5.195$ ($\Delta=+.332$, $p_{\mathrm{Holm}}=.076$). Full scored significantly above Event-Triggered ($3.059$; $d_z=3.04$) and Heuristic-Only ($1.445$; $d_z=3.65$), with both $p_{\mathrm{Holm}}<.001$. For temporal continuity, Full scored $6.474$ compared with $5.094$ for GT ($p_{\mathrm{Holm}}=.0011$). These comparisons are based entirely on human ratings and do not use RQS.

Across all 96 reactions, RQS correlated strongly with the human scores for both Pearson ($r=.855$) and Spearman ($\rho=.821$), with both $p<.05$. Because these overall correlations could partly reflect the large mean differences among the four conditions, we additionally removed the condition-level mean differences before recomputing the correlations. Pearson correlation remained $r=.908$, while Spearman correlation remained $\rho=.892$. For GT, Full, and Event-Triggered, the within-condition Spearman correlations were $.676$, $.745$, and $.888$, respectively. We therefore use RQS as a human-checked supplementary ranking metric within this evaluation setting.

The human evaluation also considers interaction-sensitive properties. Raters watched the complete speaker input and continuous listener reaction; isolated frames were not used. Contextual appropriateness and temporal continuity examine whether a reaction fits the current interaction stage and develops smoothly over time. \textit{Can Language Models Learn to Listen?} \cite{ng2023can} examines the temporal and semantic use of speaker input for listener motion, while Shirekar et al.~\cite{shirekar2025multimodal} evaluate social behaviour through synchrony, temporal alignment, and structural similarity. These directions suggest adapting speaker-listener synchrony, response timing, and trajectory alignment measures to AU or 3DMM sequences as complementary measures to human evaluation and the existing metrics.

\paragraph{Qualitative and Video Comparison.}
We align qualitative windows using the original audio clock and display gain one. The realigned C008 window spans 154.2 to 162.2s, with speaker speech occupying 54.5\% of the interval. Mean AU speed is .033345 for GT, .016843 for Full, .005424 for the output-hold control, and .000093 for Heuristic-Only. Full and output hold differ by .043779 in AU-MAE.

Three additional actual-clock windows around natural speech gaps continue to supply speaker facial observations. C001 spans 38.646 to 46.646s, C002 spans 45.078 to 53.078s, and C003 spans 70.102 to 78.102s. Fixing the emotional input to facial readout produces mean absolute AU differences from Full of .004043, .006886, and .004596. These examples visualise the emotional-state-to-facial-parameter path under the current 20-step model. Original post-rating interview feedback retains its original stimuli and model attribution.

\begin{figure*}[t]
\centering
\includegraphics[width=\linewidth]{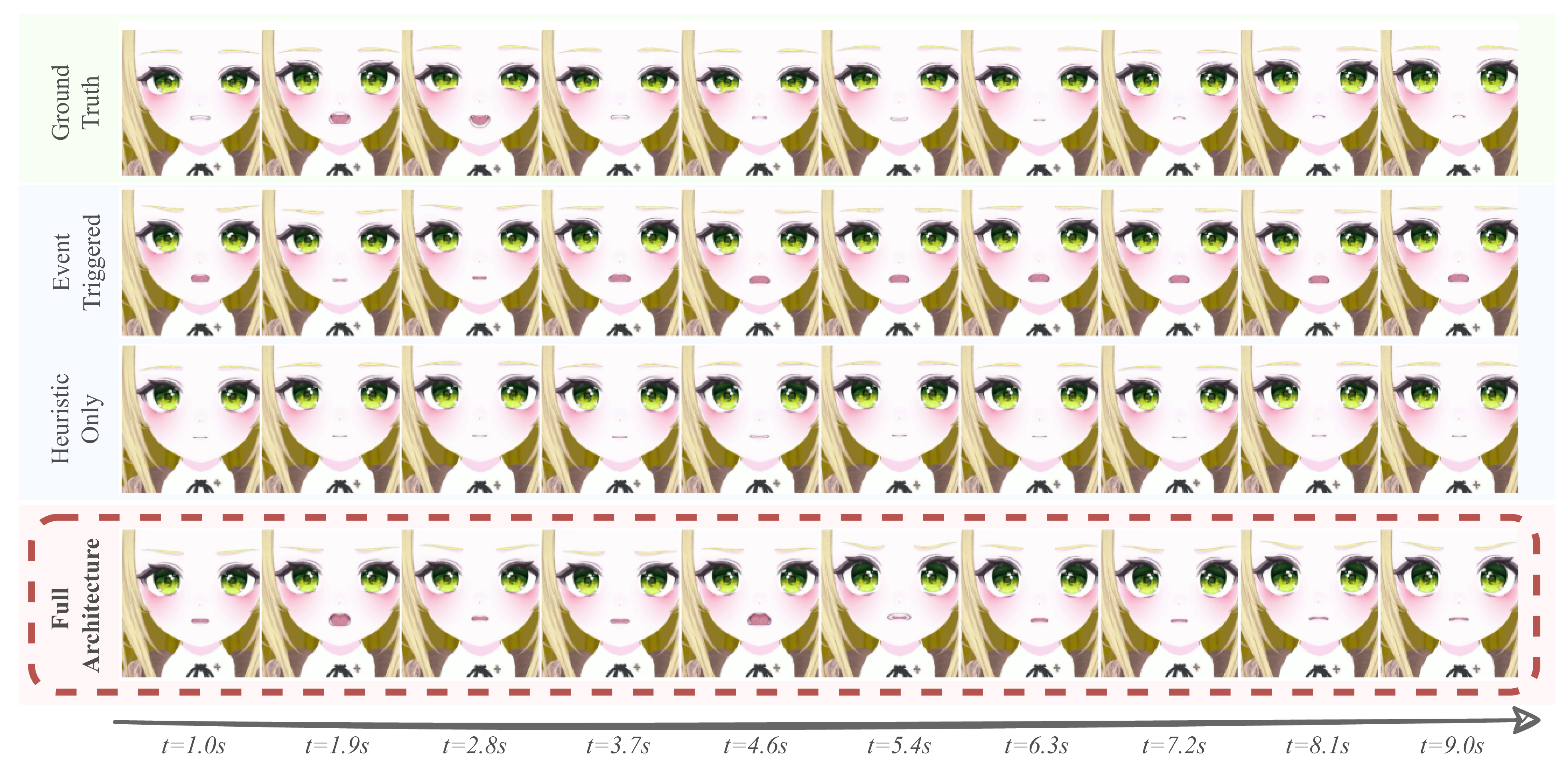}
\caption{Actual-clock trajectories for C001. Rows show GT, Full, fixed-emotion readout, and output hold. Conditions share the same eight-second time axis and display gain one. Speaker facial observations remain available during the interval.}
\label{fig:qualitative_frames}
\end{figure*}

\subsection{Comparison with Existing Automatic Metrics}
\label{app:rqs_metric_comparison}

To test whether RQS primarily reflects its own design assumptions, we compared it with FRCorr, FRDist, FRSyn, VaG, VeG, and DG on the 72 human-rated generated reactions from Full, Event-Triggered, and Heuristic-Only. Ground-truth reactions were excluded because comparing a real trajectory with itself would produce trivial self-comparison values for several trajectory-based metrics. Each generated reaction was compared with the real reaction from the same context. For lower-is-better metrics, we reversed the sign so that a positive correlation consistently indicates better agreement with human scores.

\begin{table}[htbp]
\centering
\caption{Correlation of automatic metrics with human ratings over the 72 generated reactions. Higher values indicate stronger agreement with human scores.}
\label{tab:rqs_metric_correlation}
\begin{tabular}{lcc}
\toprule
Metric & Human Pearson & Human Spearman \\
\midrule
FRCorr & -.100 & -.040 \\
$-$FRDist & .066 & .078 \\
$-$FRSyn & .000 & .000 \\
$-$VaG & .512 & .491 \\
$-$VeG & -.057 & -.001 \\
$-$DG & .189 & .136 \\
\textbf{RQS} & \textbf{.851} & \textbf{.777} \\
\bottomrule
\end{tabular}
\end{table}

RQS had the highest numerical correlations with human scores, while VaG showed a moderate raw correlation. To reduce the influence of mean differences among the three generation conditions, we centred scores within each condition. RQS remained strongly associated with human ratings, reaching $.901$ for Pearson and $.910$ for Spearman. After removing context difficulty, the correlations were $.861$ and $.881$, respectively. Within each context, RQS and human ratings agreed on condition ordering in $80.6\%$ of comparisons. These analyses use only the 72 generated reactions and are therefore distinct from the 96-stimulus human-validation analysis in Appendix~\ref{app:rqs_human}.

We next tested whether RQS provides information beyond the six automatic metrics. We used standardised ridge regression with the regularisation coefficient fixed at $1$. All conditions from the same context were held out together, while standardisation and model fitting used only the remaining contexts. MAE measures the mean absolute error between predicted and observed human scores.

\begin{table}[htbp]
\centering
\caption{Context-held-out prediction of human ratings using existing automatic metrics and RQS.}
\label{tab:rqs_incremental}
\begin{tabular}{lccc}
\toprule
Predictors & Pearson $\uparrow$ & Spearman $\uparrow$ & MAE $\downarrow$ \\
\midrule
Six automatic metrics & .480 & .432 & 1.461 \\
RQS & .847 & .716 & .851 \\
\textbf{Six metrics + RQS} & \textbf{.879} & \textbf{.786} & \textbf{.745} \\
\bottomrule
\end{tabular}
\end{table}

Adding RQS to the six automatic metrics increased both correlations with human ratings and reduced prediction error. RQS therefore provides predictive information that is not captured by FRCorr, FRDist, FRSyn, VaG, VeG, and DG alone.

Finally, we tested for residual condition-specific bias after controlling for human scores and context. Relative to Heuristic-Only, the residual difference for Full was $-.051$ ($p=.439$); relative to Event-Triggered, it was $-.129$ ($p=.0138$). Neither difference was positive. We therefore found no evidence of additional positive residual bias favouring Full under this analysis. RQS is used as a human-validated complement to existing automatic metrics rather than as a replacement for them or for human evaluation.

\section{Limitations and Societal Considerations}
\label{app:limitations}

\subsection{Interpretation of Inner Thought Flow}

Inner Thought Flow is an explicit computational text state used for subsequent generation. Controlled interventions measure its downstream effects. Human consciousness and cognition require separate evidence. The base LLM remains frozen. In the newer component procedures, automatic state hypotheses inferred from observed training input--response pairs provide retrieval memories and facial-training conditions. Their computational origin and uncertainty form part of the interpretation of these conditions.

The retained main facial DiT learns a high-frequency residual from paired-transcript conditions and true low-frequency training AUs. During inference, embeddings of thoughts and speaker words, retained V/A, and a predicted low-frequency guide populate the same interface. The sources of semantic, affective, and physical conditions therefore change between training and inference. The separately trained component models record their Action conditions and predicted-guide targets. Interpretation of each result follows the configuration used to produce it.

\subsection{Facial Interface and Rig Dependence}

The numerical generator produces 25 facial attributes, and its first 15 channels form the AU interface supplied to the avatar. A character-specific adaptor maps these values to the rig's BlendShapes. A new rig or facial representation requires the corresponding calibrated adaptor. Target-domain adaptation of the facial DiT follows the protocol in Appendix~\ref{app:seamless_adaptation}.

We evaluate calibration of controlled transformations at the final 15-dimensional AU output. The adaptor receives 15 AU values and 15 saturation indicators and predicts 15 restored values, using 465 parameters. It is fitted on 1333 official training clips, with regularisation selected on 327 internal training clips. Candidate designs use 24 development recordings. Confirmation uses 14 held-out recordings, two language seeds, three diffusion seeds, and three calibration-sampling seeds.

Across eight full-channel gain and cross-talk transformations, the maximum change in condition-mean GT AU-MAE decreases from .003466 before calibration to .000206 afterwards. AU and velocity restoration errors decrease for these transformations, with Holm-adjusted $p=.002197$ across 18 comparisons.

For 10\% cross-talk, restoration MAE decreases from .013043 to approximately zero, and GT AU-MAE changes from .199051 to .195585. Removing three AU channels gives restoration MAE .013227 before calibration and .015218 afterwards. The maximum GT AU-MAE change across all nine transformations is .003214 after calibration. These results support calibration of the specified full-channel transformations and quantify the effect of missing channel information.

We separately evaluate a V/A-only auxiliary interface. It is trained on 1471 clips, while 189 validation clips select the checkpoint. Its input is distinct from the main 72-dimensional low-frequency mapping.

On 14 original recordings aligned to the actual clock, the MAE-trained TCN obtains AU-MAE .165511, compared with .217465 for hand-crafted rules and .205370 for the training-mean control. Both comparisons have Holm-adjusted $p=.009033$ across 45 pairwise tests. The training-median control obtains .169253, with adjusted $p=1.000$ against the MAE-trained TCN.

The MAE-trained TCN has AU-CCC $-1.02\times10^{-5}$, while the hand-crafted interface has .003420. These results support learning the auxiliary interface for a specified point-error objective. Temporal correlation is reported separately through AU-CCC.

We do not treat the current rules as universal across identities. Face shape, asymmetry, timing style, and rig range require a character-specific adaptor. Related work provides additional context for this limitation. MagicFace uses AU-controlled facial expression editing while preserving identity information~\cite{wei2025magicface}. Kirchner et al.~\cite{kirchner2026controlled} study controlled AU manipulation while accounting for AU co-activation and non-target attributes. Shi et al.~\cite{shi2026capturing} explicitly model individual differences in facial expression for authentic expression generation.

\paragraph{Native Facial-Parameter Transfer.}
We evaluate a paired adaptor from recorded 25-dimensional facial attributes to the first 52 expression coefficients of the native 58-dimensional FaceVerse v2 representation. Original recordings are partitioned into 86 fitting groups, 29 development groups, and 14 disjoint confirmation groups.

The adaptor uses 55,808 paired fitting frames. ExtraTrees and .05s smoothing are selected by development MAE. Confirmation contains 588 clips from 14 recordings, includes both roles, and retains all clips with matched frame counts. Each input produces one deterministic 52-dimensional reconstruction.

\begin{table*}[htbp]
\centering
\caption{Native FaceVerse expression reconstruction on recording-disjoint confirmation data.}
\begin{tabular}{lrrr}
\toprule
Condition & Coefficient MAE & Coefficient CCC & Velocity-STD gap $\times1000$ \\
\midrule
Learnt adaptor & .067853 & .089722 & 21.64165 \\
Training mean & .074047 & .000000 & 25.67930 \\
Time-shifted command & .074223 & -.013137 & 21.47162 \\
\bottomrule
\end{tabular}
\end{table*}

Compared with the training-mean control, the adaptor reduces coefficient MAE by .006194, with a 95\% recording-bootstrap interval of [.005325, .007099]. Coefficient CCC reaches .089722. Both primary comparisons have Holm-adjusted $p=.000244$. These results support paired calibration to native expression parameters across recordings.

\subsection{Generalisation Scope}

The recording-group-disjoint REACT 2025 component study evaluates three dynamic gaps on held-out recording groups using the Action/State configuration and empirical terminal initialisation specified in Appendix~\ref{app:recording_generalisation}. Seamless Interaction evaluates target-domain adaptation with the state-update model, prompts, and low-frequency mapping fixed while the facial DiT is adapted. Reported gains compare source and adapted models under a shared mean-preserving decoder.

\subsection{Societal Considerations}

Facial reaction generation can be misused for impersonation, undisclosed synthetic interaction, emotional manipulation, and deepfakes. Practical safeguards include user notice, identity permission, generated-content labels, access control, and audit records. The setting evaluated in this work uses an openly identified virtual agent and does not imitate a real person without the user's knowledge.

\section{Sample Agent Inner Thought Prompt}
\label{app:innerthought_prompt}

The state interface supplies perceived speaker information, relevant memory, retained affect, previous thought, and the previous semantic expectation. The offline generator requests a JSON object containing four fields.

\begin{itemize}
\item \texttt{thought} contains the next first-person thought generated by the model.
\item \texttt{expectation\_error} is a reported number in $[0,1]$ comparing the previous expectation with the current observation.
\item \texttt{action} contains a proposed command or the literal string \texttt{None}.
\item \texttt{expectation} describes the anticipated immediate outcome.
\end{itemize}

The main numerical configuration uses the current progression prompt. This includes action-progression instructions under unchanged input, including progression after a None action and physical movement after three consecutive no-change waits. The newer listener-task protocols use the task-specific policy summarised below.

\begin{lstlisting}[style=promptstyle, caption={Listener-task policy summary for the controlled component protocols}]
You are the listener in a recorded seated conversation.
Use the observed speaker, available memory, current emotion,
previous thought, and previous expectation.
Training episodes describe other conversations and provide analogies.
Continuing to listen is valid. Return None when no new action is needed.
Use explicit execution feedback to determine whether an action is ongoing.
Use the previous expectation to interpret the current observation.
Keep the action and immediate expectation consistent with that reasoning.
Return a JSON object with thought, expectation_error, action, and expectation.
\end{lstlisting}

The reported error enters the following programmed affect update. Affect changes through new text appraisal, retained affect, and the error-dependent arousal and dominance terms in the emotion update described in Appendix~\ref{app:eq_validation}. The first error is set to zero when a previous expectation is unavailable.

The Action field records a proposed command. The controlled silence experiment supplies the simulated execution status defined by its protocol. Facial attributes are generated through the configured state-to-facial readout, and their 15 AU channels are sent to the rig adaptor. The generation records retain the prompt configuration and state trajectory used for each experiment.

\end{document}